\documentclass[10pt,a4paper]{article}
\usepackage[utf8]{inputenc}
\usepackage{authblk}
\usepackage{amsmath}
\usepackage{amsfonts}
\usepackage{amssymb}
\usepackage{graphicx}
\usepackage[numbers]{natbib}
\usepackage{caption} 
\usepackage{url}

\DeclareMathOperator*{\cov}{cov}

\DeclareMathOperator*{\trace}{trace}

\newcommand{\vect}[1]{\mathbf{#1}}
\renewcommand{\vec}[1]{\boldsymbol{#1}}

\author[1]{Bj\"orn Weghenkel\thanks{bjoern.weghenkel@rub.de}}
\author[2]{Asja Fischer\thanks{fischer@iro.umontreal.ca}}
\author[1]{Laurenz Wiskott\thanks{laurenz.wiskott@rub.de}}

\affil[1]{Institute for Neural Computation,
		  Ruhr-University Bochum, 
		  Bochum, 
		  Germany}
\affil[2]{Computer Science Institute\\
         University of Bonn\\
		 Bonn, 
		 Germany}

\date{}

\title{Graph-based Predictable Feature Analysis}

\begin{document}

\maketitle

\begin{center}
Preprint\\
The final publication is available at Springer via\\
\url{http://dx.doi.org/10.1007/s10994-017-5632-x}
\end{center}

\hspace{1cm}

\begin{abstract}
We propose graph-based predictable feature analysis (GPFA), a new method for unsupervised learning of predictable features from high-di\-men\-sio\-nal time series, where high predictability is understood very generically as low variance in the distribution of the next data point given the previous ones. We show how this measure of predictability can be understood in terms of graph embedding as well as how it relates to the information-theoretic measure of  predictive information in special cases. We confirm the effectiveness of GPFA on different datasets, comparing it to three existing algorithms with similar objectives---namely slow feature analysis, forecastable component analysis, and predictable feature analysis---to which GPFA shows very competitive results.
\end{abstract}

\maketitle

\section{Introduction}
\label{sec:introduction}

When we consider the problem of an agent (artificial or biological) interacting with its environment, its signal processing is naturally embedded in time. In such a scenario, a feature's ability to predict the future is a necessary condition for it to be useful in any behaviorally relevant way: A feature that does not hold information about the future is out-dated the moment it is processed and any action based on such a feature can only be expected to have random effects.

As an practical example, consider a robot interacting with its environment. When its stream of sensory input is high-dimensional (e.g., the pixel values from a camera), we are interested in mapping this input to a lower-dimensional representation to make subsequent machine learning steps and decision making more robust and efficient. At this point, however, it is crucial not to throw away information that the input stream holds about the future as any subsequent decision making will depend on this information. The same holds for time series like video, weather, or business data: When performing classification or regression on the learned features, or when the data is modelled for instance by a (hidden) Markov model, we are mostly interested in features that have some kind of predictive power.

Standard algorithms for dimensionality reduction (DR), like PCA, however, are designed to preserve properties of the data that are not (or at least not explicitly) related to predictability and thus are likely to waste valuable information that could be extracted from the data's temporal structure. In this paper we will therefore focus on the unsupervised learning of predictable features for high-dimensional time series, that is, given a sequence of data points in a high-dimensional vector space we are looking for the projection into a sub-space which makes predictions about the future most reliable.

While aspects of predictability are (implicitly) dealt with through many different approaches in machine learning, only few algorithms have addressed this problem of finding subspaces for multivariate time series suited for predicting the future. The recently proposed \emph{forecastable component analysis} (ForeCA)~\citep{Goerg-2013} is based on the idea that predictable signals can be recognized by their low entropy in the power spectrum while white noise in contrast would result in a power spectrum with maximal entropy. \emph{Predictable feature analysis} (PFA)~\citep{RichthoferWiskott-2013} focuses on signals that are well predictable through autoregressive processes. Another DR approach that was not designed to extract predictable features but explicitly takes into account the temporal structure of the data is \emph{slow feature analysis} (SFA)~\citep{WiskottSejnowski-2002}. Still, the resulting slow features can be seen as a special case of predictable features~\citep{CreutzigSprekeler-2008}. For reinforcement learning settings, \emph{predictive projections}~\citep{Sprague-2009} and \emph{robotic priors}~\citep{JonschkowskiBrock-2015} learn mappings where actions applied to similar states result in similar successor states. Also, there are recent variants of PCA that at least allow for weak statistical dependence between samples~\citep{HanLiu-2013}.

All in all, however, the field of unsupervised learning of predictable subspaces for time series is largely unexplored. Our contribution consists of a new measure of the predictability of learned features as well as of an algorithm for learning those. The proposed measure has the advantage of being very generic, of making only few assumptions about the data at hand, and of being easy to link to the information-theoretic quantity of \emph{predictive information}~\citep{BialekNemenmanEtAl-2001}, that is, the mutual information between past and future. The proposed algorithm, \emph{graph-based predictable feature analysis} (GPFA), not only shows very competitive results in practice but also has the advantage of being very flexible, and of allowing for a variety of future extensions. Through its formulation in terms of a graph embedding problem, it can be straightforwardly combined with many other, mainly geometrically motivated objectives that have been formulated in the graph embedding framework~\citep{YanXuEtAl-2007}---like Isomap \citep{TenenbaumSilvaEtAl-2000}, Locally Linear Embedding \citep[LLE, ][]{RoweisSaul-2000}, Laplacian Eigenmaps \citep{BelkinNiyogi-2003}, and Locality Preserving Projections \citep[LPP, ][]{HeNiyogi-2004}. Moreover, GPFA could make use of potential speed-ups like spectral regression~\citep{CaiHeEtAl-2007}, include additional label information in its graph like in~\citep{EscalanteWiskott-2013}, or could be applied to non-vectorial data like text. Kernelization and other approaches to use GPFA in a non-linear way are discussed in Section~\ref{sec:discussion}.

The remaining paper is structured as follows. In Section~\ref{sec:gpfa} we derive the GPFA algorithm. We start by introducing a new measure of predictability (Section~\ref{sec:measuring}), a consistent estimate for it (Section~\ref{sec:estimating}), and a simplified version of the estimate which is used by the proposed algorithm as an intermediate step (Section~\ref{sec:simplifying}). Then the link to the graph embedding framework is established in Sections~\ref{sec:pfa-graphs} and \ref{sec:graph_embedding}. After describing three useful heuristics in Section~\ref{sec:heuristics}, the core algorithm is summarized in Section~\ref{sec:algorithm} and an iterated version of the algorithm is described in Section~\ref{sec:iterated_gpfa}. Afterwards the algorithm is analyzed with respect to its objective's close relation to predictive information (Section~\ref{sec:predictive_information}) and with respect to its time complexity (Section~\ref{sec:time_complexity}). Section~\ref{sec:related_methods} summarizes the most closely related approaches for predictable feature learning---namely SFA, ForeCA, and PFA---and Section~\ref{sec:experiments} describes experiments on different datasets. We end with a discussion of limitations, open questions and ideas which shall be conducted by future research in Section~\ref{sec:discussion} and with a conclusion in Section~\ref{sec:conclusion}.

\section{Graph-based Predictable Feature Analysis}
\label{sec:gpfa}

Given is a time series $\mathbf{x}_t \in \mathbb{R}^N$, $t = 1,\dots, S$, as training data that is assumed to be generated by a stationary stochastic process $(\vec X_t)_{t}$ of order $p$. The goal of GPFA is to find a lower-dimensional feature space for that process by means of an orthogonal transformation $\vect A \in \mathbb{R}^{N \times M}$, leading to projected random variables $\vec Y_t = \vect A^T \vec X_t$ with low average variance given the state of the $p$ previous time steps. We use $\vec X_{t}^{(p)}$ to denote the concatenation $(\vec X_{t}^T, \dots, \vec X_{t-p+1}^T)^T$ of the $p$ predecessor of $\vec X_{t+1}$ to simplify notation. The corresponding state values are vectors in $\mathbb{R}^{N \cdot p}$ and denoted by $\vect x_t^{(p)}$.

\subsection{Measuring predictability}
\label{sec:measuring}

We understand the predictability of the learned feature space in terms of the variance of the projected random variables $\vec Y_t$ in this space: The lower their average variance given their $p$-step past, the higher the predictability. We measure this through the expected covariance matrix of $\vec Y_{t+1}$ given $\vec Y_{t}^{(p)}$ and minimize it in terms of its trace, i.e., we minimize the sum of variances in all principal directions. Formally, we look for the projection matrix $\vect A$ leading to a projected stochastic process $(\vec Y_t)_t$ with minimum 
\begin{equation}
\label{eq:predictability}
\trace( \mathbb{E}_{\vec Y_t^{(p)}}[ \cov( \vec Y_{t+1} | \vec Y_{t}^{(p)} ) ] ) \enspace.
\end{equation}
For simplicity, we refer to this as ``minimizing the variance'' in the following. When we make the generally reasonable assumption of $p(\vec Y_{t+1} | \vec Y_t^{(p)} = \vect y_t^{(p)})$ being Gaussian, that makes the learned features a perfect fit to be used in combination with least-squares prediction models\footnote{Note that in the Gaussian case the covariance not only covers the distribution's second moments but is sufficient to describe the higher-order moments as well.}. For non-Gaussian conditional distributions we assume the variance to function as an useful proxy for quantifying the uncertainty of the next step. Note, however, that assuming Gaussianity for the conditional distributions does not imply or require Gaussianity of $\vec X_t$ or of the joint distributions $p(\vec X_s, \vec X_t)$, $s \neq t$, which makes the predictability measure applicable to a wide range of stochastic processes.

\subsection{Estimating predictability}
\label{sec:estimating}

In practice, the expected value in~\eqref{eq:predictability} can be estimated by sampling a time series $\vect y_1, \dots, \vect y_S$ from the process $(\vec Y_{t})_t$. However, the empirical estimate for the covariance matrices $\cov( \vec Y_{t+1} | \vec Y_{t}^{(p)} = \vect y_{t}^{(p)})$, with $\vect y_{t}^{(p)} \in \mathbb{R}^{M \cdot p}$, is not directly available because there might be only one sample of $\vec Y_{t+1}^{(p)}$ with previous state value $\vect y_t^{(p)}$. Therefore we calculate a $k$-nearest neighbor (kNN) estimate instead. Intuitively, the sample size is increased by also considering the $k$ points that are most similar (e.g., in terms of Euclidean distance) to $\vect y_{t}^{(p)}$, assuming that a distribution $p(\vec Y_{t+1} | \vec Y_{t}^{(p)} = \vect{y'}_{t}^{(p)})$ is similar to $p(\vec Y_{t+1} | \vec Y_{t}^{(p)} = \vect y_{t}^{(p)})$ if $\vect{y'}_{t}^{(p)}$ is close to $\vect y_{t}^{(p)}$. In other words, we group together signals that are similar in their past $p$ steps. To that end, a set $\mathcal{K}_{t}^{(p)}$ is constructed, containing the indices of all $k$ nearest neighbors of $\vect y_{t}^{(p)}$ (plus the $0$-st neighbor, $t$ itself), i.e., $\mathcal{K}_{t}^{(p)} := \{i \: | \: \vect y_{i}^{(p)} \text{ is kNN of } \vect y_{t}^{(p)}, i = 1,\dots,S \} \cup \{ t \}$. The covariance is finally estimated based on the successors of these neighbors. Formally, the $k$-nearest neighbor estimate of \eqref{eq:predictability} is given by
\begin{equation}
\label{eq:knn_estimate}
\trace( \langle \cov( \{ \vect y_{i+1} \: | \: i \in \mathcal{K}_{t}^{(p)}\} ) \rangle_t )\enspace,
\end{equation}
where $\langle \cdot \rangle_t$ denotes the average over $t$. Note that the distance measure used for the $k$ nearest neighbors does not necessarily need to be Euclidean. Think for instance of ``perceived similarities'' of words or faces.

While we introduce the kNN estimate here to assess the uncertainty inherent in the stochastic process, we note that it may be of practical use in a deterministic setting as well. For a deterministic dynamical system the kNN estimate includes nearby points belonging to nearby trajectories in the dataset. Thus, the resulting feature space may be understood as one with small divergence of neighboring trajectories (as measured through the Lyapunov exponent, for instance).

\subsection{Simplifying predictability}
\label{sec:simplifying}

Finding the transformation $\vect A$ that leads to the most predictable $(\vec Y_t)_{t}$ in the sense of \eqref{eq:predictability} becomes difficult through the circumstance that the predictability can only be evaluated \emph{after} $\vect A$ has been fixed. The circular nature of this optimization problem motivates the iterated algorithm described in Section~\ref{sec:iterated_gpfa}. As a helpful intermediate step we define a weaker measure of predictability that is conditioned on the input $\vec X_t$ instead of the features $\vec Y_t$ and has a closed-form solution, namely minimizing
\begin{equation*}
\trace( \mathbb{E}_{\vec X_t^{(p)}}[ \cov( \vec Y_{t+1} | \vec X_{t}^{(p)} ) ] )
\end{equation*}
via its $k$-nearest neighbor estimate
\begin{equation}
\label{eq:predictability_estimated_weak}
\trace( \langle \cov( \{ \vect y_{i+1} \: | \: i \in \tilde{\mathcal{K}}_t^{(p)}\} ) \rangle_t )\enspace.
\end{equation}
Analogous to $\mathcal{K}_t^{(p)}$, the set $\tilde{\mathcal{K}}_t^{(p)}$ contains the indices of the $k$ nearest neighbors of $\vect x_{t}^{(p)}$ plus $t$ itself. Under certain mild mixing assumptions for the stochastic process, the text-book results on $k$-nearest neighbor estimates can be applied to auto-regressive time series as well \citep{Collomb-1985}. Thus, in the limit of $S \rightarrow \infty$, $k \rightarrow \infty$, $k/S \rightarrow 0$, the estimated covariance
\begin{equation*}
\cov(\{ \vect y_{i+1} \: | \: i \in \tilde{\mathcal{K}}_t^{(p)}\}) 
= \langle \vect y_{i+1} \vect y_{i+1}^T \rangle_{i \in \tilde{\mathcal{K}}_t^{(p)}} - \langle \vect y_{i+1} \rangle_{i \in \tilde{\mathcal{K}}_t^{(p)}} \langle \vect y_{i+1} \rangle_{i \in \tilde{\mathcal{K}}_t^{(p)}}^T
\end{equation*}
converges to 
\begin{equation*}
\mathbb{E}[\vec Y_{t+1} \vec Y_{t+1}^T | \vec X_t^{(p)} = \vect x_t^{(p)}] - \mathbb{E}[\vec Y_{t+1} | \vec X_t^{(p)} = \vect x_t^{(p)}] \mathbb{E}[\vec Y_{t+1} | \vec X_t^{(p)} = \vect x_t^{(p)}]^T \enspace,
\end{equation*}
i.e., it is a consistent estimator of $\cov(\vec Y_{t+1}| \vec X_t^{(p)} = \vect x_t^{(p)})$.

When measuring predictability, one assumption made about the process $(\vec X_t)_{t}$ in the following is that it is already white, i.e., $\mathbb{E}[\vec X_t] = \vect 0$ and $\cov(\vec X_t) = \vect I$ for all $t$. Otherwise components with lower variance would tend to have higher predictability \emph{per se}.

\subsection{Predictability as graph}
\label{sec:pfa-graphs}

Instead of optimizing objective~\eqref{eq:predictability_estimated_weak} directly, we reformulate it such that it can be interpreted as the embedding of an undirected graph on the set of training samples. Consider the graph to be represented by a symmetric connection matrix $\mathbf{W} = (\mathrm{W}_{ij})_{ij}\in \mathbb{R}^{S \times S}$ with weights $\mathrm{W}_{ij} = \mathrm{W}_{ji} > 0$ whenever two nodes corresponding to vectors $\mathbf{x}_i$ and $\mathbf{x}_j$ from the training sequence are connected by an edge $\{ \mathbf{x}_i, \mathbf{x}_j \}$. Further assume an orthogonal transformation $\mathbf{A} \in \mathbb{R}^{N \times M}$ for that graph with $M \ll N$ that minimizes
\begin{equation}
\label{eq:LPP}
\sum_{i,j=1}^{S} {\mathrm{W}_{ij} \|\mathbf{A}^T \mathbf x_i - \mathbf{A}^T \mathbf x_j\|^2}=\sum_{i,j=1}^{S} {\mathrm{W}_{ij} \|\mathbf y_i - \mathbf y_j\|^2} \enspace.
\end{equation}
Intuitively, this term becomes small if the projections of points connected in the graph (i.e., nodes for which $\mathrm W_{ij} > 0$) are close to each other, while there is no penalty for placing the projections of unconnected points far apart.

Through a proper selection of the weights $\mathrm{W}_{ij}$, the transformation $\vect A$ can be used to maximize predictability in the sense of minimizing \eqref{eq:predictability_estimated_weak}. This becomes clear by noting that the trace of the sample covariance 
\begin{align*}
\cov( \{ \vect y_{i+1} \: | \: i \in \tilde{\mathcal{K}}_t^{(p)}\} ) 
&= \langle \vect y_{i+1} \vect y_{i+1}^T \rangle_{i \in \tilde{\mathcal{K}}_t^{(p)}} - \langle \vect y_{i+1} \rangle_{i \in \tilde{\mathcal{K}}_t^{(p)}} \langle \vect y_{i+1} \rangle_{i \in \tilde{\mathcal{K}}_t^{(p)}}^T\\
&= \langle (\vect y_{i+1} - \overline{\vect y_{i+1}}) (\vect y_{i+1} - \overline{\vect y_{i+1}})^T \rangle_{i \in \tilde{\mathcal{K}}_t^{(p)}} \enspace,
\end{align*}
with $\overline{\vect y_{i+1}} = \langle \vect y_{i+1} \rangle_{i \in \tilde{\mathcal{K}}_t^{(p)}}$ being the sample mean, can always be formulated via pairwise differences of samples, since
\begin{multline}
\label{eq:sample_variance_pairwise}
\trace(\langle (\vect y_{i+1} - \overline{\vect y_{i+1}}) (\vect y_{i+1} - \overline{\vect y_{i+1}})^T \rangle_{i \in \tilde{\mathcal{K}}_t^{(p)}}) \\
\begin{aligned}
&= \langle (\vect y_{i+1} - \overline{\vect y_{i+1}})^T (\vect y_{i+1} - \overline{\vect y_{i+1}}) \rangle_{i \in \tilde{\mathcal{K}}_t^{(p)}} \\
&= \langle \vect y_{i+1}^T \vect y_{i+1} \rangle_{i \in \tilde{\mathcal{K}}_t^{(p)}} - \langle \vect y_{i+1} \rangle^T_{i \in \tilde{\mathcal{K}}_t^{(p)}} \langle \vect y_{j+1} \rangle_{j \in \tilde{\mathcal{K}}_t^{(p)}} \\
&= \langle \vect y_{i+1}^T \vect y_{i+1} - \vect y_{i+1}^T \vect y_{j+1} \rangle_{i,j \in \tilde{\mathcal{K}}_t^{(p)}} \\
&= \frac{1}{2} \langle \vect y_{i+1}^T \vect y_{i+1} - 2 \vect y_{i+1}^T \vect y_{j+1} + \vect y_{j+1}^T \vect y_{j+1} \rangle_{i,j \in \tilde{\mathcal{K}}_t^{(p)}} \\
&= \frac{1}{2} \langle \| \vect y_{i+1} - \vect y_{j+1} \|^2 \rangle_{i,j \in \tilde{\mathcal{K}}_t^{(p)}} \enspace.
\end{aligned}
\end{multline}
Thus, by incrementing weights\footnote{All edge weights are initialized with zero.} of the edges $\{\vect y_{i+1}, \vect y_{j+1}\}$ for all $i,j \in \tilde{\mathcal{K}}_t^{(p)}$, $t = p, \dots, S-1$, minimizing \eqref{eq:LPP} directly leads to the minimization of \eqref{eq:predictability_estimated_weak}.

Note that for the construction of the graph, the data actually does not need to be represented by points in a vector space. Data points also could, for instance, be words from a text corpus as long as there are either enough samples per word or there is an applicable distance measure to determine ``neighboring words'' for the $k$-nearest neighbor estimates.

\subsection{Graph embedding}
\label{sec:graph_embedding}

To find the orthogonal transformation $\mathbf{A} = (\mathbf{a}_1, \mathbf{a}_2, \dots, \mathbf{a}_M) \in  \mathbb{R}^{N \times M}$ that minimizes \eqref{eq:LPP}, let the training data be concatenated in $\vect X = (\mathbf{x}_1, \mathbf{x}_2, \dots, \mathbf{x}_S) \in  \mathbb{R}^{N \times S}$, and let $\mathbf{D} \in \mathbb{R}^{S \times S}$ be a diagonal matrix with $\mathrm{D}_{ii} = \sum_{j}\mathrm{W}_{ij}$ being the sum of edge weights connected to node $\mathbf x_i$. Let further $\vect L := \vect D - \vect W$ be the graph Laplacian. Then, the minimization of \eqref{eq:LPP} can be re-formulated as a minimization of 
\begin{multline}
\label{eq:lpp_derivation}
\frac{1}{2} \sum_{i,j=1}^{S} {\mathrm{W}_{ij} \| \mathbf{A}^T \mathbf x_i - \mathbf{A}^T \mathbf x_j\|^2} \\
\enspace\enspace\enspace\enspace = \frac{1}{2} \sum_{i,j=1}^{S} {\mathrm{W}_{ij} \trace( ( \mathbf{A}^T \mathbf x_i - \mathbf{A}^T \mathbf x_j ) ( \mathbf{A}^T \mathbf x_i - \mathbf{A}^T \mathbf x_j )^T )} \\
= \trace( \sum_{i=1}^{S} \mathbf{A}^T \mathbf x_i \mathrm D_{ii} \mathbf x_i^T \mathbf{A}  - \sum_{i,j=1}^{S} \mathbf{A}^T \mathbf x_i \mathrm W_{ij} \mathbf x_j^T \mathbf{A} ) \\
= \trace( \mathbf{A}^T \mathbf X( \mathbf D- \mathbf W )  \mathbf X^T \mathbf A )
= \trace( \mathbf{A}^T \mathbf X \mathbf L \mathbf X^T \mathbf A ) 
= \sum_{i=1}^M \vect a_i^T \vect X \vect L \vect X^T \vect a_i \enspace.
\end{multline}
The $\vect a_i$ that minimize \eqref{eq:lpp_derivation} are given by the first (``smallest'') $M$ eigenvectors of the eigenvalue problem
\begin{equation}
\label{eq:eigenproblem}
\vect X \vect L \vect X^T {\vect a} = \lambda \vect a \enspace.
\end{equation}
See \citep{HeNiyogi-2004} for the analogous derivation of the one-dimensional case that was largely adopted here as well as for a kernelized version of the graph embedding.

\subsection{Additional heuristics}
\label{sec:heuristics}

The following three heuristics proved to be useful for improving the results in practice.

\subsubsection*{Normalized graph embedding}

First, in the context of graph embedding, the minimization of $\vect a^T \vect X \vect L \vect X^T \vect a$ described in the section above is often solved subject to the additional constraint $\vect a^T\vect X \vect D \vect X^T \vect a = 1$ (see for instance~\citep{HeNiyogi-2004,VonLuxburg2007}). Through this constraint the projected data points are normalized with respect to their degree of connectivity in every component $\vect Y = \vect X^T \vect a$, i.e., $\vect Y^T \vect D \vect Y = 1$. Objective function and constraint can be combined in the Lagrange function $\vect a^T \mathbf X \mathbf L \mathbf X^T \vect a - \lambda (\vect a^T\vect X \vect D \vect X^T \vect a - 1)$. Then the solution is given by the  ``smallest'' eigenvectors of the generalized eigenvalue problem 
\begin{equation}
\label{eq:generalized_eigenproblem}
\vect X \vect L \vect X^T {\vect a} = \lambda \vect X \vect D \vect X^T {\vect a} \enspace.
\end{equation}
 Solving this generalized eigenvalue problem instead of \eqref{eq:eigenproblem} tended to improve the results for GPFA.

\subsubsection*{Minimizing variance of the past}

Second, while not being directly linked to the above measure of predictability, results benefit significantly when the variance of the past is minimized simultaneously to that of the future. To be precise, additional edges $\{\vect y_{i-p}, \vect y_{j-p}\}$ are added to the graph for all $i,j \in \tilde{\mathcal{K}}_t^{(p)}$, $t=p+1,\dots,S$. The proposed edges here have the effect of mapping states with \emph{similar} futures to \emph{similar} locations in feature space. In other words, states are represented with respect to what is expected in the next steps (not with respect to their past). Conceptually this is related to the idea of \emph{causal states}~\citep{ShaliziCrutchfield-2001}, where all (discrete) states that share the same conditional distribution over possible futures are mapped to the same causal state (also see~\citep{Still-2009} for a closely related formulation in interactive settings).

\subsubsection*{Star-like graph structure}

As a third heuristic, the graph above can be simplified by replacing the sample mean $\overline{\vect y_{i+1}}$ in the minimization objective
\begin{multline*}
\trace( \langle \cov( \{ \vect y_{i+1} \: | \: i \in \tilde{\mathcal{K}}_t^{(p)}\} ) \rangle_t ) \\
= \trace(\langle (\vect y_{i+1} - \overline{\vect y_{i+1}}) (\vect y_{i+1} - \overline{\vect y_{i+1}})^T \rangle_{i \in \tilde{\mathcal{K}}_t^{(p)}})  \\
= \langle \| \vect y_{i+1} - \overline{\vect y_{i+1}} \|^2 \rangle_{i \in \tilde{\mathcal{K}}_t^{(p)}}
\end{multline*}
by $\vect y_{t+1}$. This leads to
\begin{equation}
\label{eq:graph2}
\langle \| \vect y_{i+1} - \vect y_{t+1} \|^2 \rangle_{i \in \tilde{\mathcal{K}}_t^{(p)}} \enspace,
\end{equation}
inducing a graph with star-like structures. It is constructed by adding (undirected) edges $\{\vect y_{i+1}, \vect y_{t+1}\}$ for all $i \in \tilde{\mathcal{K}}_t^{(p)}$. Analogously, edges for reducing the variance of the past are given by $\{\vect y_{i-p}, \vect y_{t-p}\}$ for $i \in \tilde{\mathcal{K}}_t^{(p)}$.

\begin{figure}[htb]
\centering
  \includegraphics[width=.4\textwidth]{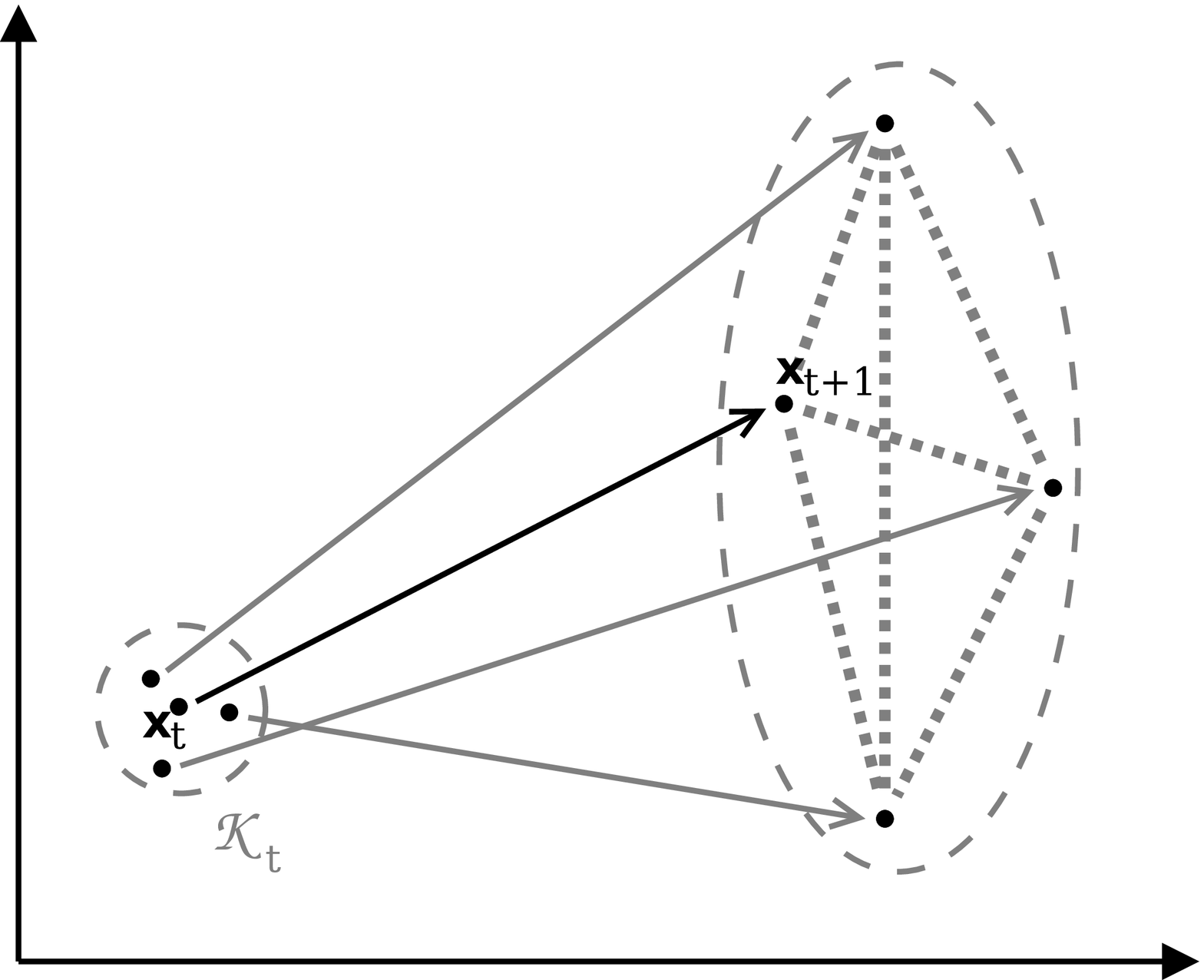}
  \hspace{.5cm}
  \includegraphics[width=.4\textwidth]{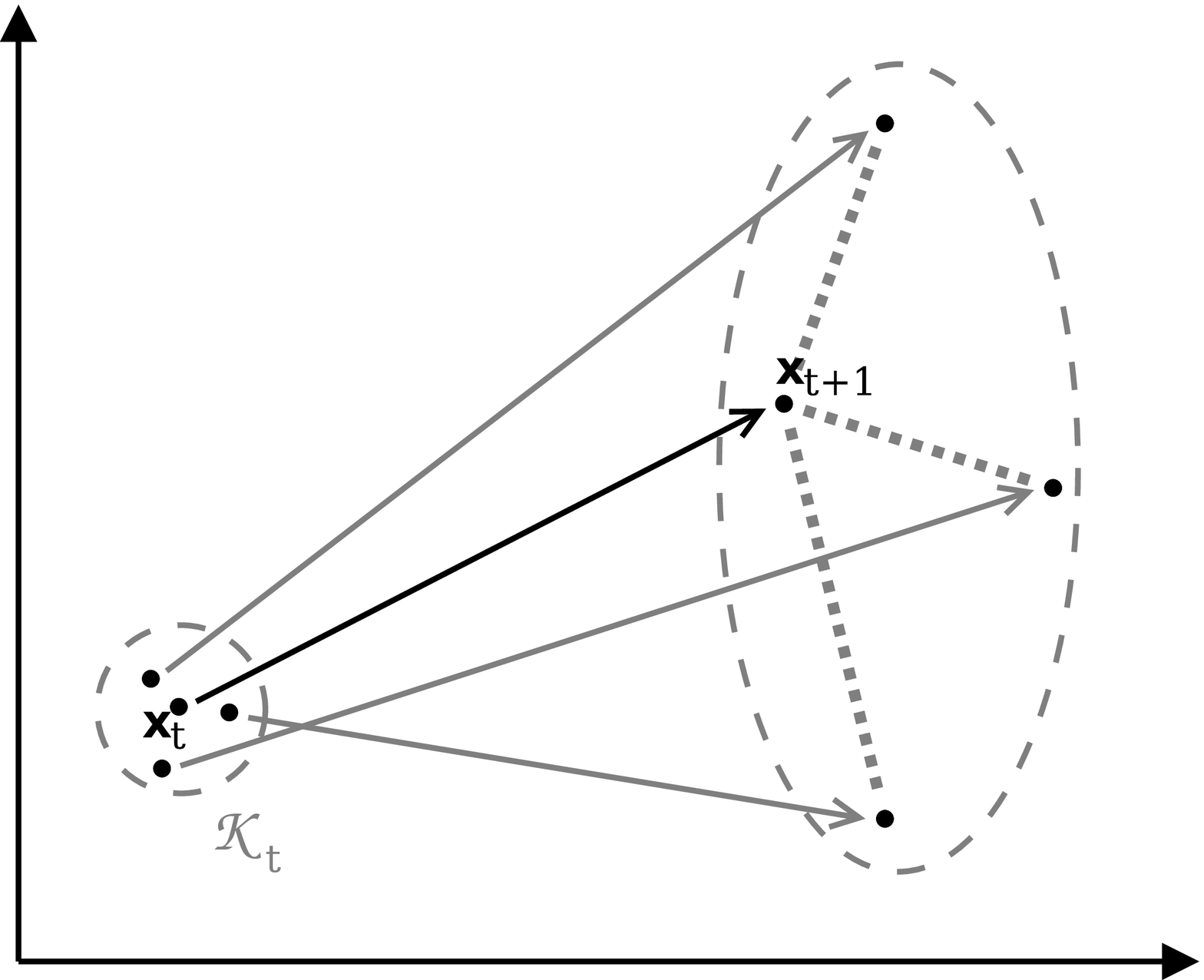}
\caption{Graphs constructed for GPFA (1) and GPFA (2) are illustrated on the left and right, respectively. Both pictures show a sample $\vect x_t$ and its $k$ nearest neighbors together with their successors in time (indicated through arrows). The distribution of the successors indicates that the first axis can be predicted with less uncertainty than the second axis. The dotted lines depict edges that are added to the graph according to the two variants of the algorithm. Edges for minimizing the variance of the past are constructed analogously.}
\label{fig:graph_construction}
\end{figure}

We refer to the resulting algorithms as GPFA (1) and GPFA (2), corresponding to the graphs defined through \eqref{eq:sample_variance_pairwise} and \eqref{eq:graph2}, respectively. See Figure~\ref{fig:graph_construction} for an illustration of both graphs. The differences in performance are empirically evaluated in Section~\ref{sec:experiments}.

\subsection{Algorithm}
\label{sec:algorithm}

In the following, the core algorithm is summarized step by step, where training data $\vect x_1, \dots, \vect x_S$ is assumed to be white already or preprocessed accordingly (in that case, the same transformation has to be taken into account during subsequent feature extractions). Lines starting with (1) and (2) indicate the steps for GPFA~(1) and GPFA~(2), respectively.

\begin{enumerate}

\item \textbf{Calculate neighborhood}

For every $\mathbf{x}_t^{(p)}$, $t=p,\dots, S$, calculate index set $\tilde{\mathcal{K}}_t^{(p)}$ of $k$ nearest neighbors (plus $t$ itself).

\item \textbf{Construct graph (future)}

Initialize connection matrix $\mathbf W$ to zero. For every $t=p,\dots, S-1$, add edges, according to either

\begin{enumerate}

\item[({1})] $\mathrm W_{i+1, j+1} \leftarrow \mathrm W_{i+1, j+1} + 1 \; \forall i,j \in \tilde{\mathcal{K}}_t^{(p)}$ or

\item[({2})] $\mathrm W_{i+1, t+1} \leftarrow \mathrm W_{i+1, t+1} + 1$ and \\
$\mathrm W_{t+1, i+1} \leftarrow \mathrm W_{t+1, i+1} + 1 \; \forall i \in \tilde{\mathcal{K}}_t^{(p)} \setminus \{t\}$.

\end{enumerate}

\item \textbf{Construct graph (past)}

For every $t=p+1,\dots,S$, add edges, according to either

\begin{enumerate}

\item[({1})] $\mathrm W_{i-p, j-p} \leftarrow \mathrm W_{i-p, j-p} + 1 \; \forall i,j \in \tilde{\mathcal{K}}_t^{(p)}$ or

\item[({2})] $\mathrm W_{i-p, t-p} \leftarrow \mathrm W_{i-p, t-p} + 1$ and \\
$\mathrm W_{t-p, i-p} \leftarrow \mathrm W_{t-p, i-p} + 1 \; \forall i \in \tilde{\mathcal{K}}_t^{(p)} \setminus \{t\}$.

\end{enumerate}

\item \textbf{Linear graph embedding}

Calculate $\vect L$ and $\vect D$ as defined in Section \ref{sec:graph_embedding}.

Find the first (``smallest'') $M$ solutions to $\mathbf{X} \mathbf{L} \mathbf{X}^T \mathbf{a} = \lambda \mathbf{X} \mathbf{D} \mathbf{X}^T \mathbf{a}$ and normalize them, i.e., $\|\vect a\| = 1$.

\end{enumerate}

\subsection{Iterated GPFA}
\label{sec:iterated_gpfa}

\begin{figure}[htb]
\centering
\includegraphics[width=0.4\textwidth]{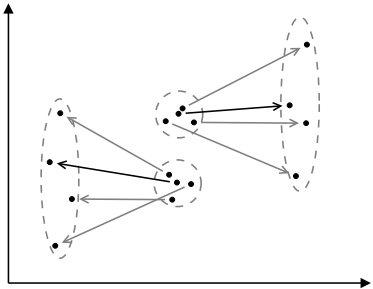}
\caption{Schematic illustration of a case where $p(\vect y_{t+1} | \vec X_t  = \vect x_t)$ and $p(\vect y_{t+1}| \vec Y_t = \vect y_t)$ differ significantly. Points from two neighborhoods are shown together with their immediate successors in time. The distributions of the successors indicate that the first axis would be the most predictable direction. However, projecting all points on the first axis would result in a feature that is highly unpredictable. Therefore another direction will likely be preferred in the next iteration.}
\label{fig:pfa_problem}
\end{figure}

As shown in Section \ref{sec:pfa-graphs}, the core algorithm above produces features $(\vec Y_t)_t$ with low $\trace( \mathbb{E}_{\vec X_t^{(p)}}[ \cov(\vec Y_{t+1} | \vec X_t^{(p)} ) ] )$. In many cases these features may already be predictable in themselves, that is, they have a low  $\trace( \mathbb{E}_{\vec Y_t^{(p)}}[ \cov(\vec Y_{t+1} | \vec Y_t^{(p)} ) ] )$. There are, however, cases where the results of both objectives can differ significantly (see Figure~\ref{fig:pfa_problem} for an example of such a case). Also, the $k$-nearest neighbor estimates of the covariances become increasingly unreliable in higher-dimensional spaces.

Therefore, we propose an iterated version of the core algorithm as a heuristic to address these problems. First, an approximation of the desired covariances $\cov(\vec Y_{t+1} | \vec Y_t^{(p)} = \vect y_t^{(p)})$ can be achieved by rebuilding the graph according to neighbors of $\vect y_t^{(p)}$, not $\vect x_t^{(p)}$. This in turn may change the whole optimization problem, which is the reason to repeat the whole procedure several times. Second, calculating the sample covariance matrices based on the $k$ nearest neighbors of $\vect y_t^{(p)} \in \mathbb{R}^{M \cdot p}$ instead of $\vect x_t^{(p)} \in \mathbb{R}^{N \cdot p}$ counteracts the problem of unreliable $k$-nearest neighbor estimates in high-dimensional spaces, since $M \cdot p \ll N \cdot p$.

The resulting (iterated) GPFA algorithm works like this:

\begin{itemize}

\item[a)] Calculate neighborhoods $\tilde{\mathcal{K}}_t^{(p)}$ of $\vect x_t^{(p)}$ for $t=p, \dots, S-1$.

\item[b)] Perform steps 2--4 of GPFA as described in Section\,\ref{sec:algorithm}.

\item[c)] Calculate projections $\vect y_t = \vect A^T \vect x_t$ for $t=1, \dots, S$.

\item[d)] Calculate neighborhoods\footnote{Of course, this step is not necessary for the last iteration.} $\mathcal{K}_t^{(p)}$ of $\vect y_t^{(p)}$ for $t=p, \dots, S-1$.

\item[e)] Start from step b), using $\mathcal{K}_t^{(p)}$ instead of $\tilde{\mathcal{K}}_t^{(p)}$.
\end{itemize}
where steps b) to e) are either repeated for $R$ iterations or until convergence.

While we can not provide a theoretical guarantee for the iterative process to converge, it did so in practice in all of our experiments  (see Section~\ref{sec:experiments}). Also note that in general there is no need for the dimensionality $M$ of the intermediate projections $\vect y_t \in \mathbb{R}^M$ to be the same as for the final feature space. 

\subsection{Relationship to predictive information}
\label{sec:predictive_information}

\emph{Predictive information}---that is, the mutual information between past states and future states---has been used as a natural measure of how well-predictable a stochastic process is (e.g., \citep{BialekTishby-1999} and \citep{ShaliziCrutchfield-2001}). In this section we discuss under which conditions the objective of GPFA corresponds to extracting features with maximal predictive information.

Consider again the stationary stochastic process $(\vec X_t)_{t}$ of order $p$ and its extracted features $\vec Y_t = \vect A^T \vec X_t$. Their predictive information is given by
\begin{equation}
\label{eq:predictive_information}
\vec I(\vec Y_{t+1} ; \vec  Y_t^{(p)}) = H(\vec Y_{t+1}) - H(\vec Y_{t+1} | \vec Y_t^{(p)}) \enspace,
\end{equation} 
where $H(\vec Y_{t+1})= \mathbb{E}[-\log p( \vec Y_{t+1} )]$ denotes the entropy and 
\begin{equation*}
H(\vec Y_{t+1} | \vec Y_t^{(p)}) = 
\mathbb{E}_{\vec Y_{t+1}, \vec Y_t^{(p)}} [-\log p(\vec Y_{t+1} | \vec Y_t^{(p)})]
\end{equation*}
denotes the conditional entropy of $(\vec Y_{t+1})_{t}$ given its past.

If we assume $\vec Y_{t+1}$ to be normally distributed---which can be justified by the fact that it corresponds to a mixture of a potentially high number of distributions from the original high-dimensional space---then its differential entropy is given by $H( \vec Y_{t+1} ) = \frac{1}{2} \log \{ (2 \pi e)^M \} + \log \{ | \cov( \vec Y_{t+1} ) | \}$ and is thus a strictly increasing function of the determinant of its covariance. Now recall that $(\vec X_t)_{t}$ is assumed to have zero mean and covariance $\vect I$. Thus, $\cov(\vec Y_{t+1}) = \vect I$ holds independently of the selected transformation $\vect A$ which makes $H(\vec Y_{t+1})$ independent of $\vect A$ too.

What remains for the maximization of~\eqref{eq:predictive_information} is the minimization of the term $H(\vec Y_{t+1} | \vec Y_t^{(p)})$. Again assuming Gaussian distributions, the differential conditional entropy is given by 
\begin{equation}
\label{eq:conditional_entropy}
\begin{split}
H( \vec Y_{t+1} | \vec Y_t^{(p)} ) &= \frac{1}{2} \log \{ (2 \pi e)^M \} \\
&+ \mathbb{E}_{\vec Y_t^{(p)}} [ \log \{ | \cov( \vec Y_{t+1} | \vec Y_t^{(p)} ) | \} ] \enspace.
\end{split}
\end{equation}
When we consider the special case of the conditional covariance $\cov( \vec Y_{t+1} | \vec Y_t^{(p)} = \vect y_t^{(p)} ) =: \vect \Sigma_{\vec Y_{t+1} | \vec Y_t^{(p)}}$ being the same for every value that $\vec Y_t^{(p)}$ may take, then the expected value reduces to $\log \{ | \vect \Sigma_{\vec Y_{t+1} | \vec Y_t^{(p) } } | \}$ and \eqref{eq:conditional_entropy} becomes minimal for the projection for which the resulting determinant $| \vect \Sigma_{\vec Y_{t+1} | \vec Y_t^{(p) } } |$ is minimized. Furthermore, under this assumption, \eqref{eq:predictability} can be written as
\begin{multline*}
\trace( \mathbb{E}_{\vec Y_t^{(p)}} [ \cov( \vec Y_{t+1} | \vec Y_t^{(p)} ) ] ) \\
= \trace( \vect \Sigma_{\vec Y_{t+1} | \vec Y_t^{(p)}} ) \\
= \trace( \vect A^T \vect \Sigma_{\vec X_{t+1} | \vec Y_t^{(p)}} \vect A ) \enspace.
\end{multline*}
Thus, it becomes clear that $\vect \Sigma_{\vec Y_{t+1} | \vec Y_t^{(p)}}$ with minimal trace is constructed by selecting the principle directions from the $N \times N$ matrix $\vect \Sigma_{\vec X_{t+1} | \vec Y_t^{(p)}}$ that correspond to the $M$ smallest eigenvalues. Thereby the determinant $| \vect \Sigma_{\vec Y_{t+1} | \vec Y_t^{(p)}} |$ is minimized as well, since---like for the trace---its minimization only depends on the selection of the smallest eigenvalues. Thus, GPFA produces features with the maximum predictive information under this assumption of a prediction error $\vect \Sigma_{\vec Y_{t+1} | \vec Y_t^{(p)}}$ independent of the value of $\vec Y_t^{(p)}$ (and to the degree that the iterated heuristic in Section~\ref{sec:iterated_gpfa} minimizes~\eqref{eq:predictability}).

For the general case of different $\cov( \vec Y_{t+1} | \vec Y_t^{(p)} )$ for different values of $\vec Y_t^{(p)}$ we have the following equality for the last term in~\eqref{eq:conditional_entropy}:
\begin{multline}
\label{eq:det_to_trace}
\mathbb{E}_{\vec Y_t^{(p)}} [ \log \{ | \cov( \vec Y_{t+1} | \vec Y_t^{(p)} ) | \} ]\\
= \mathbb{E}_{\vec Y_t^{(p)}} [ \trace( \log \{ \cov( \vec Y_{t+1} | \vec Y_t^{(p)} ) \} ) ]\\
= \trace( \mathbb{E}_{\vec Y_t^{(p)}} [ \log \{ \cov( \vec Y_{t+1} | \vec Y_t^{(p)} ) \} ] )\enspace.
\end{multline}
This corresponds to GPFA's objective~\eqref{eq:predictability} with logarithmically weighted covariances. Such a weighting intuitively makes sense from the perspective that the predictive information expresses how many bits of uncertainty (that is, variance) are removed through knowing about the feature's past. Since the number of bits only grows logarithmically with increasing variance, the weight of events with low uncertainty is disproportionally large, which could be accounted for in the objective function if the goal would be low coding length instead of low future variance.

\subsection{Time complexity}
\label{sec:time_complexity}

In the following section we derive GPFA's asymptotic time complexity in dependence of the number of training samples $S$, input dimensions $N$, process order $p$, output dimensions $M$, number of iterations $R$, as well as the neighborhood size $k$.

\subsubsection{$k$-nearest-neighbor search}

The first computationally expensive step of the GPFA is the $k$-nearest-neighbor search. When we naively assume a brute-force approach, it can be realized in $\mathcal{O}(N p S)$. This search is repeated for each of the $S$ data points and for each of the $R$ iterations (in $N \cdot p$ dimensions for the first iteration and in $M \cdot p$ for all others). Thus, the $k$-nearest-neighbor search in the worst case has a time complexity of
\begin{equation*}
\mathcal{O}(N p S^2 + R M p S^2) \enspace.
\end{equation*}
Of course, more efficient approaches to $k$-nearest-neighbor search exist.

\subsubsection{Matrix multiplications}
\label{sec:time_complexity_matrix_mult}

The second expensive step consists of the matrix multiplications in \eqref{eq:generalized_eigenproblem} to calculate the projected graph Laplacians. For a multiplication of two dense matrices of size $l \times m$ and $m \times n$ we assume a computational cost of $\mathcal{O}(l m n)$. If the first matrix is sparse, with $L$ being the number of non-zero elements, we assume $\mathcal{O}(L n)$. This gives us a complexity of $\mathcal{O}(N^2 S + L N)$ for the left-hand side of~\eqref{eq:generalized_eigenproblem}. For GPFA~(1) there is a maximum of $L = 2 k^2 S$ non-zero elements (corresponding to the edges added to the graph, which are not all unique), for GPFA~(2) there is a maximum of $L = 2 k S$. The right-hand side of~\eqref{eq:generalized_eigenproblem} then can be ignored since it's complexity of $\mathcal{O}(N^2 S + S N)$ is completely dominated by the left-hand side. Factoring in the number of iterations $R$, we finally have computational costs of 
\begin{equation*}
\mathcal{O}(R N^2 S + R L N)
\end{equation*}
with $L = k^2 S$ for GPFA~(1) and $L = k S$ for GPFA~(2).

\subsubsection{Eigenvalue decomposition}

For solving the eigenvalue problem \eqref{eq:generalized_eigenproblem} $R$ times we assume an additional time complexity of $\mathcal{O}(R N^3)$. This is again a conservative guess because only the first $M$ eigenvectors need to be calculated.

\subsubsection{Overall time complexity}

\sloppy{Taking together the components above, GPFA has a time complexity of} $\mathcal{O}(N p S^2 + R M p S^2 + R N^2 S + R L N + R N^3)$ with $L = k^2 S$ for GPFA~(1) and $L = k S$ for GPFA~(2). In terms of the individual variables, that is: $\mathcal{O}(S^2)$, $\mathcal{O}(N^3)$, $\mathcal{O}(M)$, $\mathcal{O}(p)$, $\mathcal{O}(R)$, and $\mathcal{O}(k^2)$ or $\mathcal{O}(k)$ for GPFA~(1) or GPFA~(2), respectively.

\section{Related methods}
\label{sec:related_methods}

In this section we briefly summarize the algorithms most closely related to GPFA, namely SFA, ForeCA, and PFA.

\subsection{SFA}

\begin{sloppypar}
Although SFA originally has been developed to model aspects of the visual cortex, it has been successfully applied to different problems in technical domains as well (see \citep{Escalante-B.Wiskott-2012a} for a short overview), like, for example, state-of-the art age-estimation \citep{EscalanteWiskott-2016}. It is one of the few DR algorithms that considers the temporal structure of the data. In particular, slowly varying signals can be seen as a special case of predictable features \citep{CreutzigSprekeler-2008}. It is also possible to reformulate the slowness principle implemented by SFA in terms of graph embedding, for instance to incorporate label information into the optimization problem \citep{EscalanteWiskott-2013}.
\end{sloppypar}

Adopting the notation from above, SFA finds an orthogonal transformation $\vect A \in \mathbb{R}^{N \times M}$ such that the extracted signals $\vect y_t = \vect A^T \vect x_t$ have minimum temporal variation $\langle \| \vect y_{t+1} - \vect y_t \|^2 \rangle_t$. The input vectors $\vect x_t$---and thus $\vect y_t$ as well---are assumed to be white.

\subsection{ForeCA}

In case of ForeCA~\citep{Goerg-2013}, $(\vec X_t)_{t}$ is assumed to be a stationary second-order process and the goal of the algorithm is finding an extraction vector $\vect a$ such that the projected signals $Y_t = \vect a^T \vec X_t$ are as \emph{forecastable} as possible, that is, having a low entropy in their power spectrum. Like SFA, ForeCA has the advantage of being completely model- and parameter-free. 

For the formal definition of \emph{forecastability}, first consider the signal's autocovariance function $\gamma_Y(l) = E(Y_t - \mu_Y) E(Y_{t-l} - \mu_Y)$, with $\mu_Y$ being the mean value and the corresponding autocorrelation function $\rho_Y(l) = \gamma_Y(l) / \gamma_Y(0)$. The spectral density of the process can be calculated as the  Fourier transform of the autocorrelation function, i.e., as
\begin{equation*}
f_Y(\lambda) = \sum_{j=-\infty}^{\infty} \rho_Y(j)e^{ij\lambda} \enspace,
\end{equation*}
with $i = \sqrt{-1}$ being the imaginary unit.

Since $f_Y(\lambda) \geq 0$ and $\int_{-\pi}^\pi f_Y(\lambda)d\lambda = 1$, the spectral density can be interpreted as a probability density function and thus its entropy calculated as
\begin{equation*}
H(Y_t) = - \int_{-\pi}^{\pi} f_Y(\lambda) \log(f_Y(\lambda)) d\lambda \enspace.
\end{equation*}
For white noise the spectral density becomes uniform with entropy $\log(2\pi)$.
This motivates the definition of \emph{forecastability} as
\begin{equation*}
\Omega(Y_t) := 1 - \frac{H(Y_t)}{\log(2\pi)} \enspace,
\end{equation*}
with values between $0$ (white noise) and $\infty$ (most predictable). 
Since $\Omega(Y_t)=\Omega(\vect a^T \vec X_t$) is invariant to scaling and shifting, $\vec X_t$ can be assumed to be white, without loss of generality. The resulting optimization problem
\begin{equation*}
\textrm{arg\,max}_\vect a \Omega(\vect a^T \vec X_t)
\end{equation*}
then is solved by an EM-like algorithm that uses weighted overlapping segment averaging (WOSA) to estimate the spectral density of a given (training) time series. By subsequently finding projections which are orthogonal to the already extracted ones, the approach can be employed for finding projections to higher dimensional subspaces as well. For details about ForeCA see \citep{Goerg-2013}.

\subsection{PFA}

The motivation behind PFA is finding an orthogonal transformation $\vect A \in \mathbb{R}^{N \times M}$ as well as coefficient matrices $\vect B_i \in \mathbb{R}^{M \times M}$, with $i = 1 \dots p$, such that the linear, autoregressive prediction error of order $p$,
\begin{equation*}
\langle \| \vect A^T \vect x_t - \sum_{i=1}^p \vect B_i \vect A^T \vect x_{t-i} \|^2 \rangle_t \enspace,
\end{equation*}
is minimized. However, this is a difficult problem to optimize because the optimal values of $\vect A$ and $\vect B_i$ mutually depend on each other. Therefore the solution is approached via a related but easier optimization problem: Let $\vect \zeta_t := (\vect x_{t-1}^T, \dots, \vect x_{t-p}^T)^T \in \mathbb{R}^{N \cdot p}$ be a vector containing the $p$-step history of $\vect x_t$. Let further $\vect W \in \mathbb{R}^{N \times N \cdot p}$ contain the coefficients that minimize the error of predicting $\vect x_t$ from its own history, i.e., $\langle \| \vect x_t - \vect W \vect \zeta_t \|^2 \rangle_t$. Then minimizing $\langle \| \vect A^T \vect x_t - \vect A^T \vect W \vect \zeta_t \|^2 \rangle_t$ with respect to $\vect A$ corresponds to a PCA (in the sense of finding the directions of smallest variance) on that prediction error. Minimizing this prediction error however does not necessarily lead to features $\vect y_t = \vect A^T \vect x_t$ that are best for predicting their own future because the calculated prediction was based on the history of $\vect x_t$, not $\vect y_t$ alone. Therefore an additional heuristic is proposed that is based on the intuition that the inherited errors of $K$ times repeated autoregressive predictions create an even stronger incentive to avoid  unpredictable components. Finally,
\begin{equation*}
\sum_{i=0}^K \langle \| \vect A^T \vect x_t - \vect A^T \vect W \vect V^i \vect \zeta_t \|^2 \rangle_t
\end{equation*}
is minimized with respect to $\vect A$, where $\vect V \in \mathbb{R}^{N \cdot p \times N \cdot p}$ contains the coefficients that minimize the prediction error $\langle \| \vect \zeta_{t+1} - \vect V \vect \zeta_t \|^2 \rangle_t$. 

Like the other algorithms, PFA includes a preprocessing step to whiten the data. So far, PFA has been shown to work on artificially generated data. For further details about PFA see \citep{RichthoferWiskott-2013}.

\section{Experiments}
\label{sec:experiments}

We conducted experiments\footnote{GPFA and experiments have been implemented in Python 2.7. Code and datasets will be published upon acceptance.} on different datasets to compare GPFA to SFA, ForeCA, and PFA. As a baseline, we compared the features extracted by all algorithms to features that were created by projecting into an arbitrary (i.e., randomly selected) $M$-dimensional subspace of the data's $N$-dimensional vector space.
 
For all experiments, first the training set was whitened and then the same whitening transformation was applied to the test set. After training, the learned projection was used to extract the most predictable $M$-dimensional signal from the test set with each of the algorithms. The extracted signals were evaluated in terms of their empirical predictability~\eqref{eq:knn_estimate}. The neighborhood size used for this evaluation is called $q$ in the following to distinguish it from the neighborhood size $k$ used during the training of GPFA. Since there is no natural choice for the different evaluation functions that effectively result from different $q$, we arbitrarily chose $q=10$ but also include plots on how results change with the value of $q$. The size of training and test set will be denoted by $S_{train}$ and $S_{test}$, respectively. The plots show mean and standard deviation for $50$ repetitions of each experiment.\footnote{Note that while the algorithms themselves do not depend on any random effects, the data set generation does.}

\subsection{Toy example (``predictable noise'')}
\label{sec:experiments:toy}

We created a small toy data set to demonstrate performance differences of the different algorithms. The data set contains a particular kind of predictable signals which are challenging to identify for most algorithms. Furthermore, the example is suited to get an impression for running time constants of the different algorithms that are not apparent from the big $\mathcal{O}$ notation in Section~\ref{sec:time_complexity}.

First, a two-dimensional signal
\begin{equation}
\vect x_t = \left( \genfrac{}{}{0pt}{}{\xi_t}{\xi_{t-1}} \right)
\end{equation}
was generated with $\xi_t$ being normally distributed noise. Half of the variance in this sequence can be predicted when $\vect x_{t-1}$ is known (i.e., $p=1$), making the noise partly predictable. This two-dimensional signal was augmented with $N-2$ additional dimensions of normally distributed noise to create the full data set. We generated such data sets with up to $S_{train} = 800$ training samples, a fixed test set size of $S_{test} = 100$, and with up to $N=100$ input dimensions and extracted $M=2$ components with each of the algorithms. If not varied themselves during the experiment, values were fixed to $S_{train} = 700$ training samples, $N=10$ input dimensions, and $k=10$ neighbors for the training of GPFA. The results of PFA did not change significantly with number of iterations $K$, which was therefore set to $K=0$.

\begin{figure}[htbp]
\centering
\includegraphics[width=.32\textwidth]{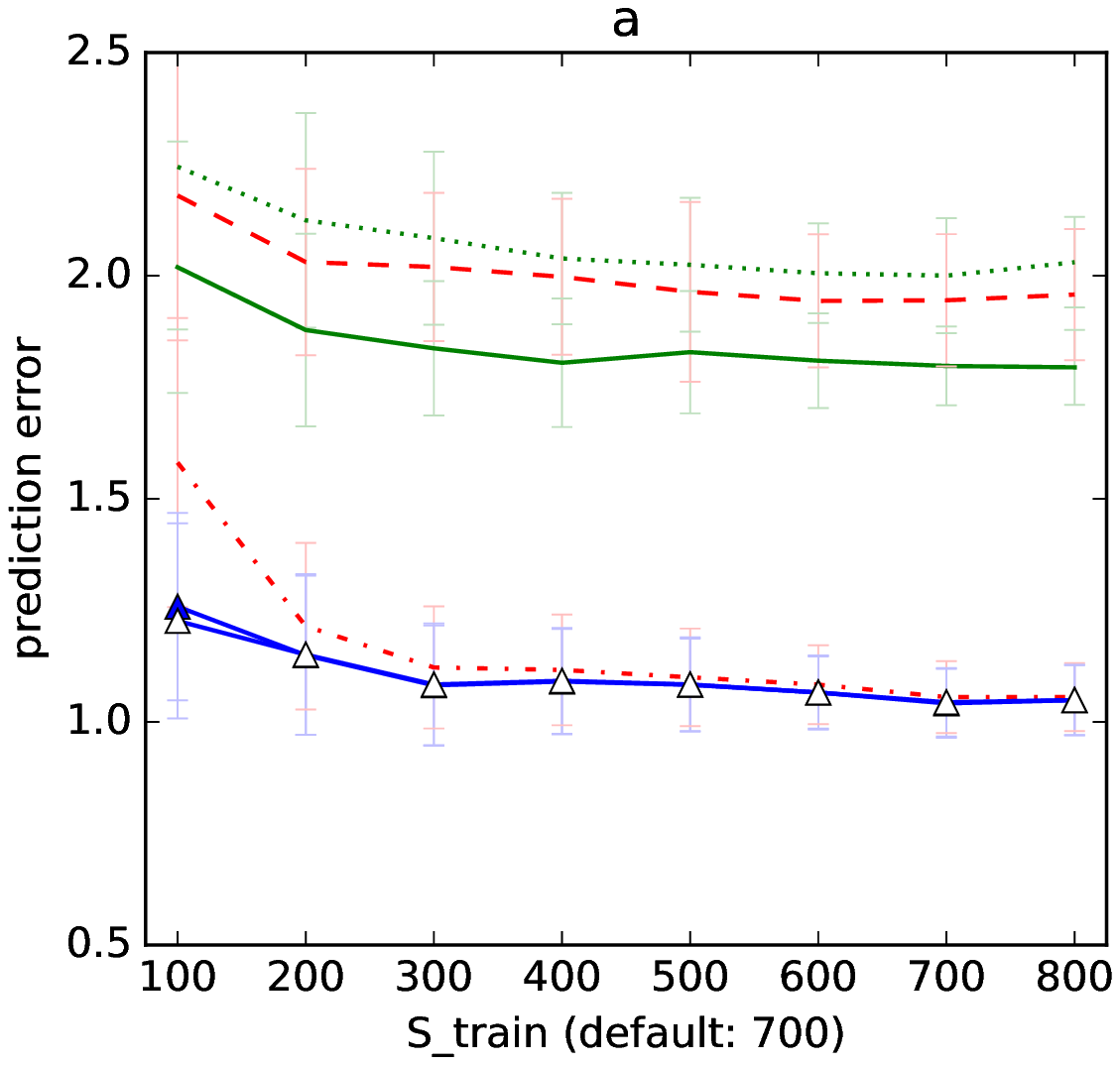}
\includegraphics[width=.32\textwidth]{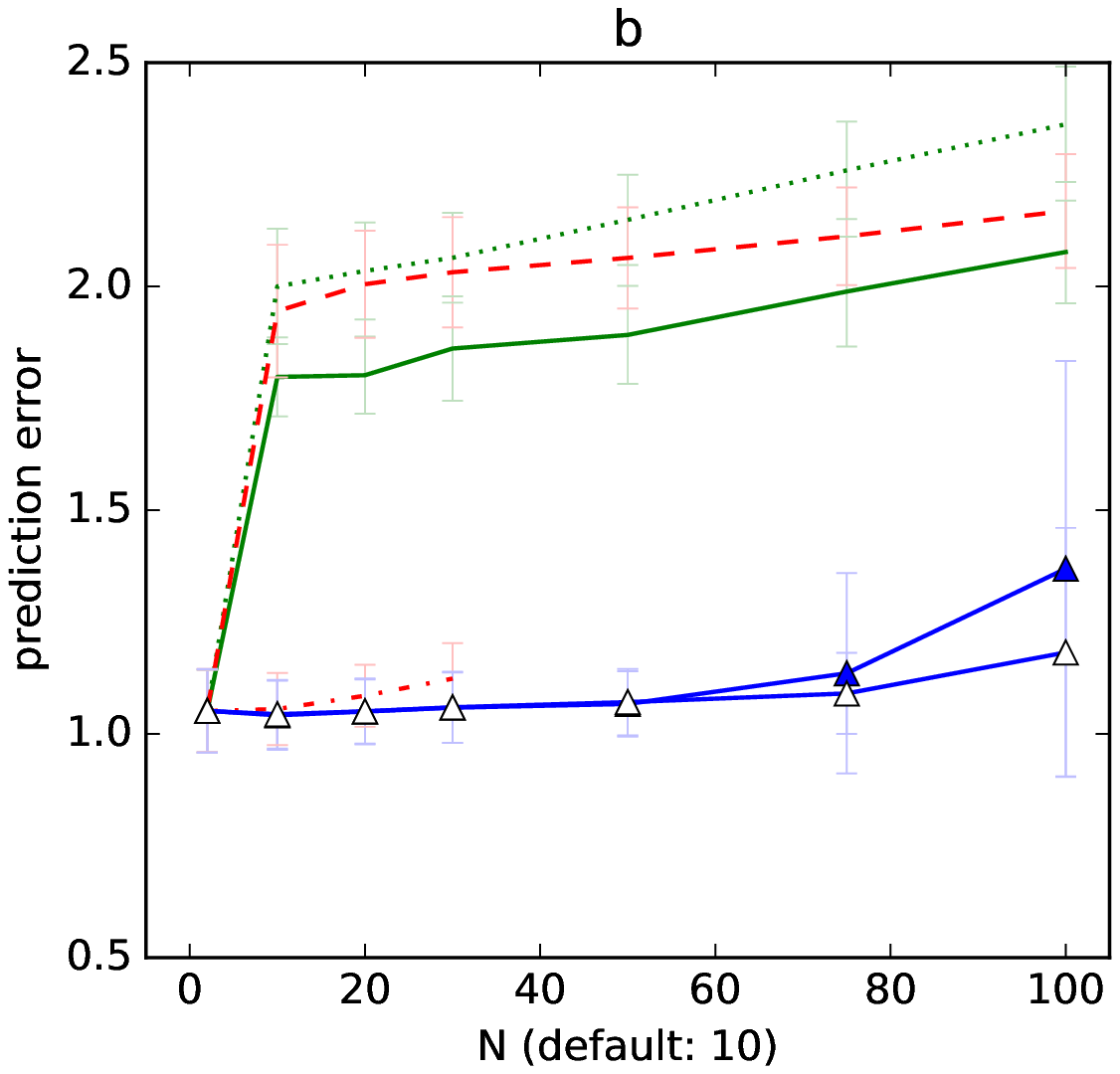}
\includegraphics[width=.32\textwidth]{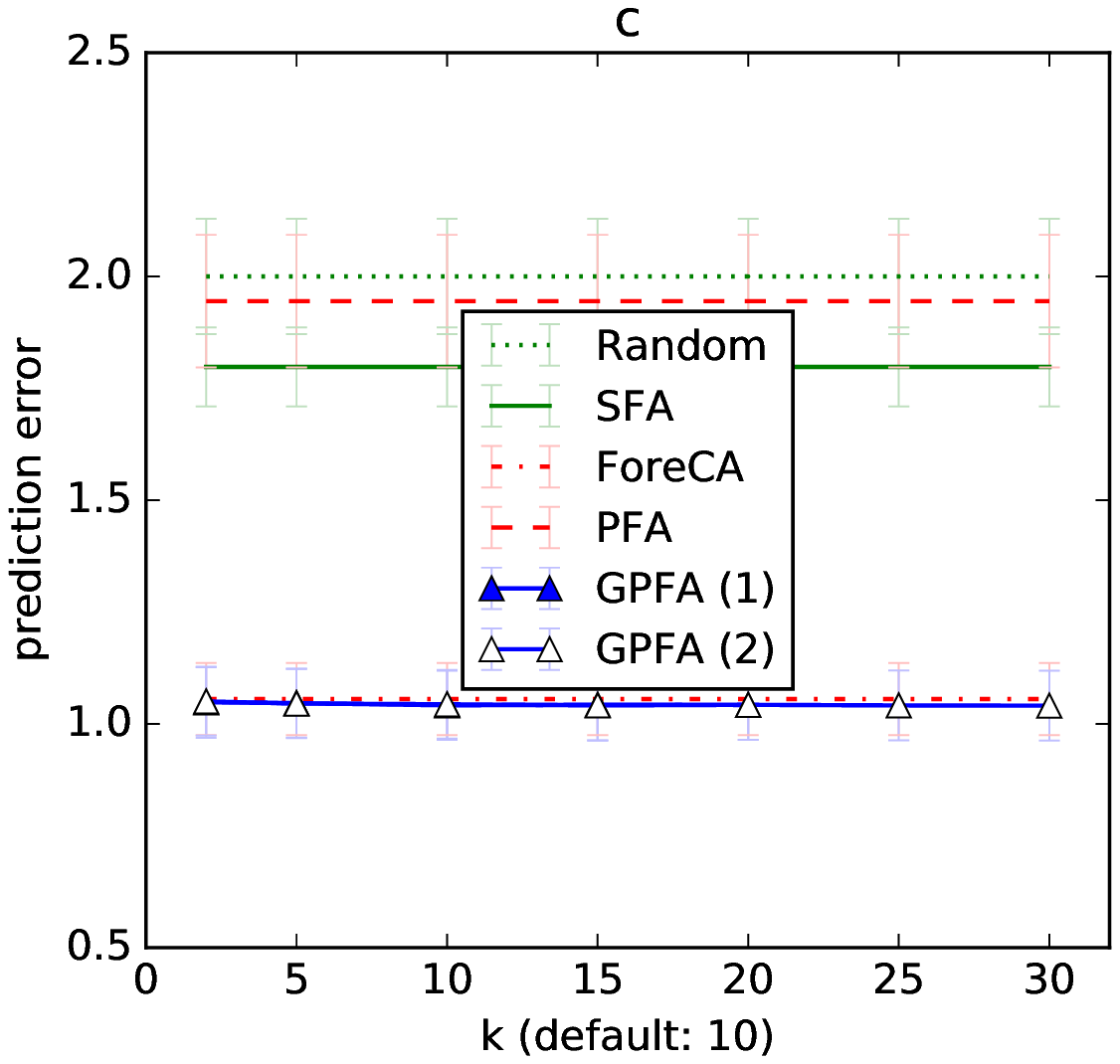}
\caption{Predictability in terms of \eqref{eq:knn_estimate} of two dimensional signals ($M=2$) extracted from the the toy dataset by the different algorithms. If not varied during the experiment, parameters were $p=1$, $k=10$, $q=10$, $S_{train}=700$, $S_{test}=100$, $N=10$, $R=50$, and $K=0$.}
\label{fig:results_noise}
\end{figure}

Figure~\ref{fig:results_noise} shows the predictability of the signals extracted by the different algorithms and how it varies in $S_{train}$, $N$, and $k$. Only ForeCA and GPFA are able to distinguish the two components of predictable noise from the unpredictable ones, as can be seen from reaching a variance of about $1$, which corresponds to the variance of the two generated, partly predictable components. As Figure~\ref{fig:results_noise}b shows, the performance of both versions of GPFA (as of all other algorithms) declines with a higher number of input dimensions (but for GPFA~(2) less than for GPFA~(1)). At this point, a larger number of training samples is necessary to produce more reliable results (experiments not shown). The results do not differ much with the choice of $k$ though.

\begin{figure}[htbp]
\centering
\includegraphics[width=.32\textwidth]{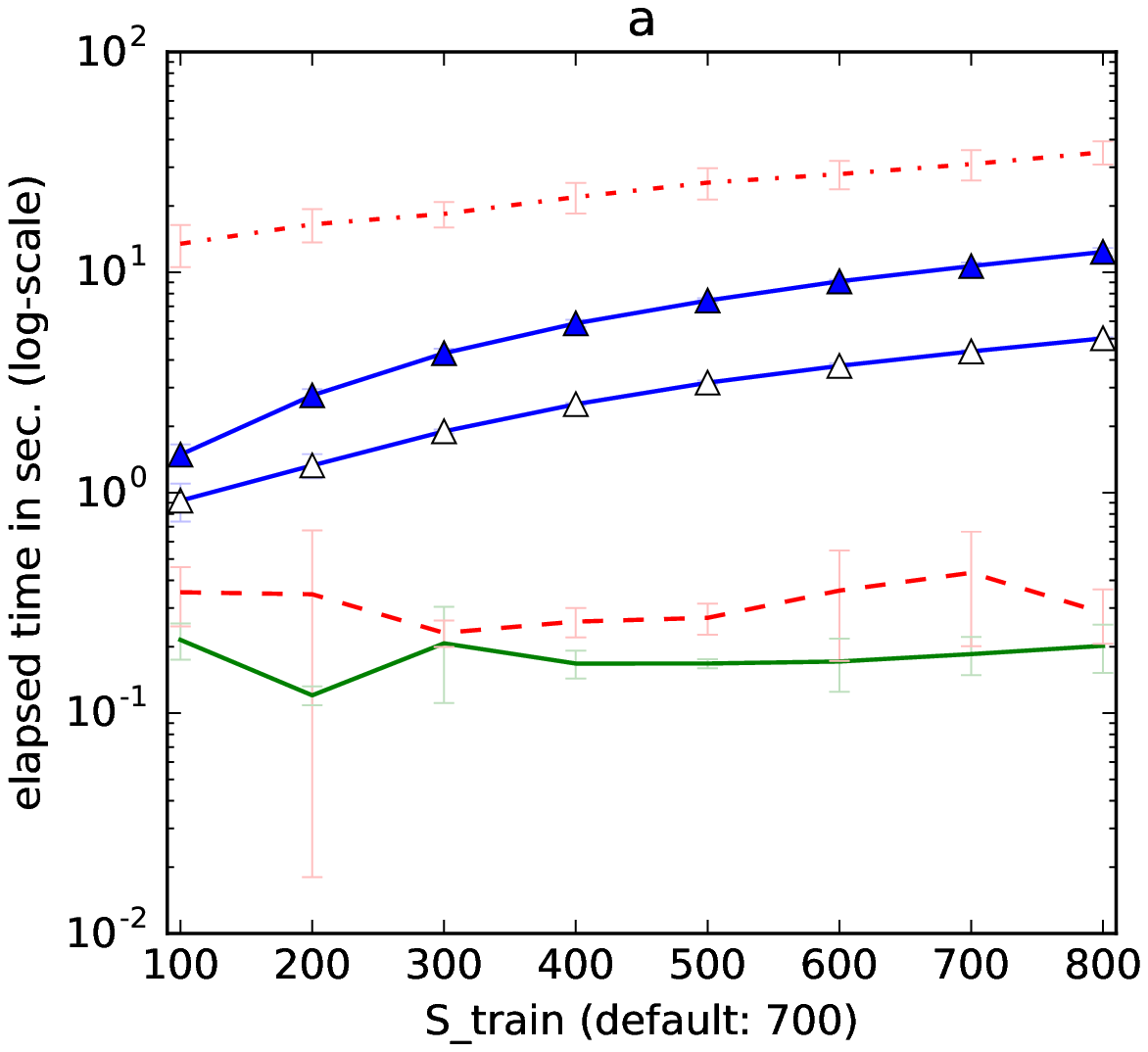}
\includegraphics[width=.32\textwidth]{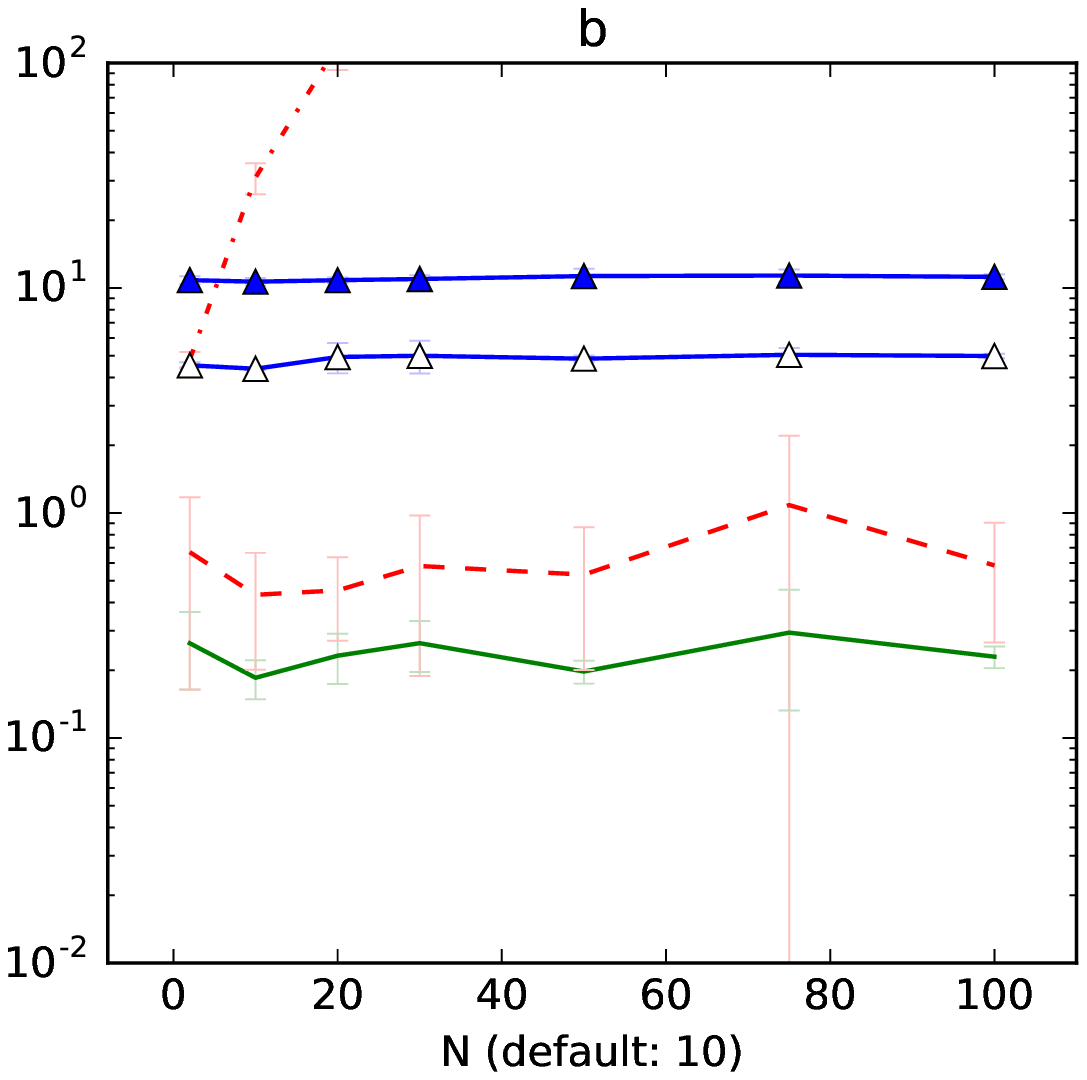}
\includegraphics[width=.32\textwidth]{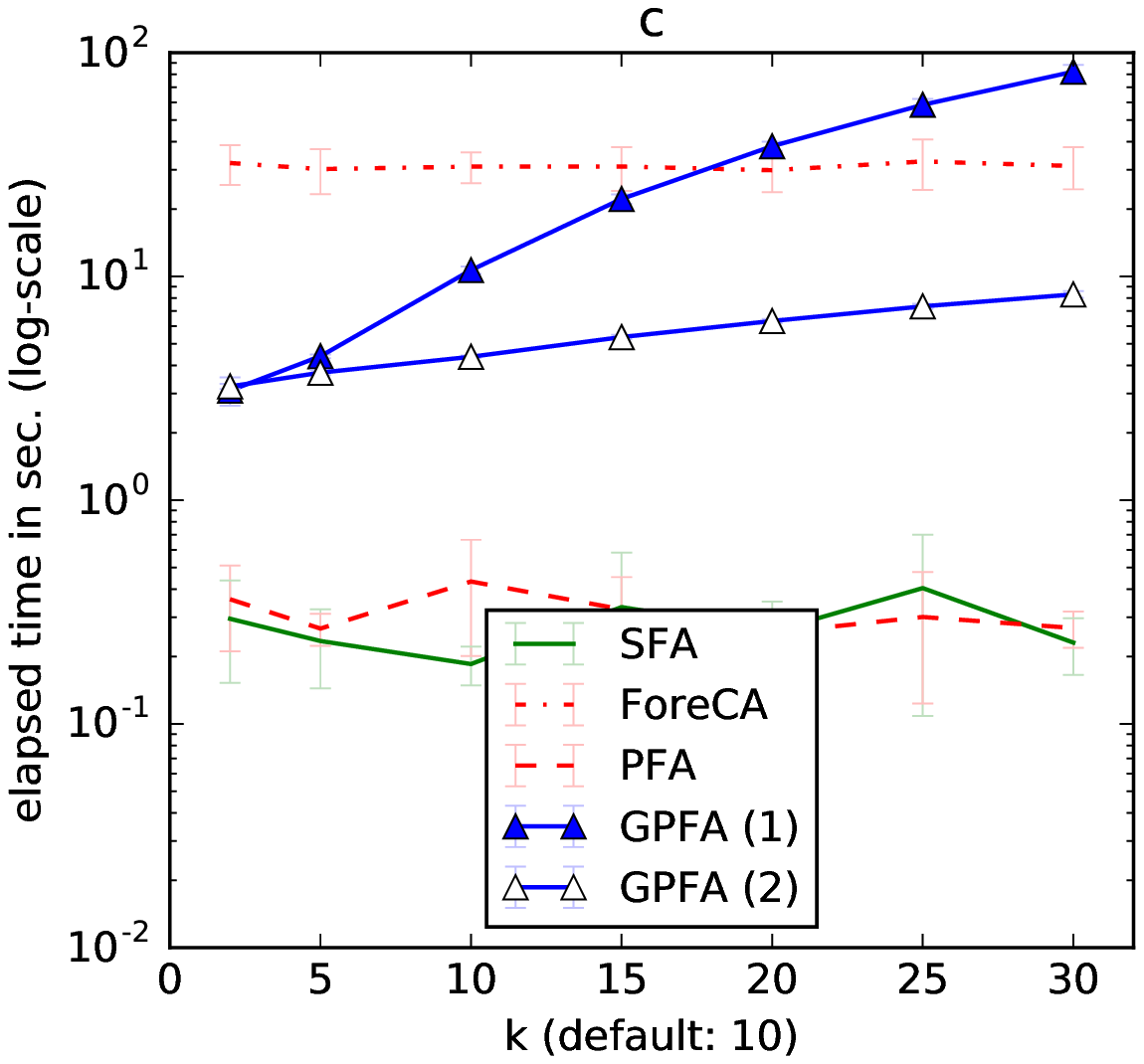}
\caption{Runtime for the experiments in Figure~\ref{fig:results_noise}.}
\label{fig:runtime_noise}
\end{figure}

As the runtime plots of the experiments reveal (see Figure~\ref{fig:runtime_noise}), ForeCA scales especially badly in the number of input dimensions $N$, so that it becomes very computationally expensive to be applied to time series with more than a few dozen dimensions. For that reason we excluded ForeCA from the remaining, high-dimensional experiments.

\subsection{Auditory data}

In the second set of experiments we focused on short-time Fourier transforms (STFTs) of audio files.
Three public domain audio files (a silent film piano soundtrack, ambient sounds from a bar, and ambient sounds from a forest) were re-sampled to $22$kHz mono. The STFTs were calculated with the Python library \texttt{stft} with a frame length of 512 and a cosine window function, resulting in three datasets with $26147$, $27427$, and $70433$ frames, respectively, each with $512$ dimensions (after discarding complex-conjugates and representing the remaining complex values as two real values each). For each repetition of the experiment, $S_{train} = 10000$ successive frames were picked randomly as training set and $S_{test} = 5000$ distinct and successive frames were picked as test set. PCA was calculated for each training set to preserve $99\%$ of the variance and this transformation was applied to training and test set alike.

The critical parameters $p$ and $k$, defining the assumed order of the process and the neighborhood size respectively, were selected through cross-validation to be a good compromise between working well for all values of $M$ and also not treating one of the algorithms unfavourably. PFA and GPFA tend to benefit from the same values for $p$. The number of iteration $R$ for GPFA was found to be not very critical and was set to $R=50$. The iteration parameter $K$ of PFA was selected by searching for the best result in $\{0\dots10\}$, leaving all other parameters fixed. 

\begin{figure}[htb]
\centering
\includegraphics[width=.45\textwidth]{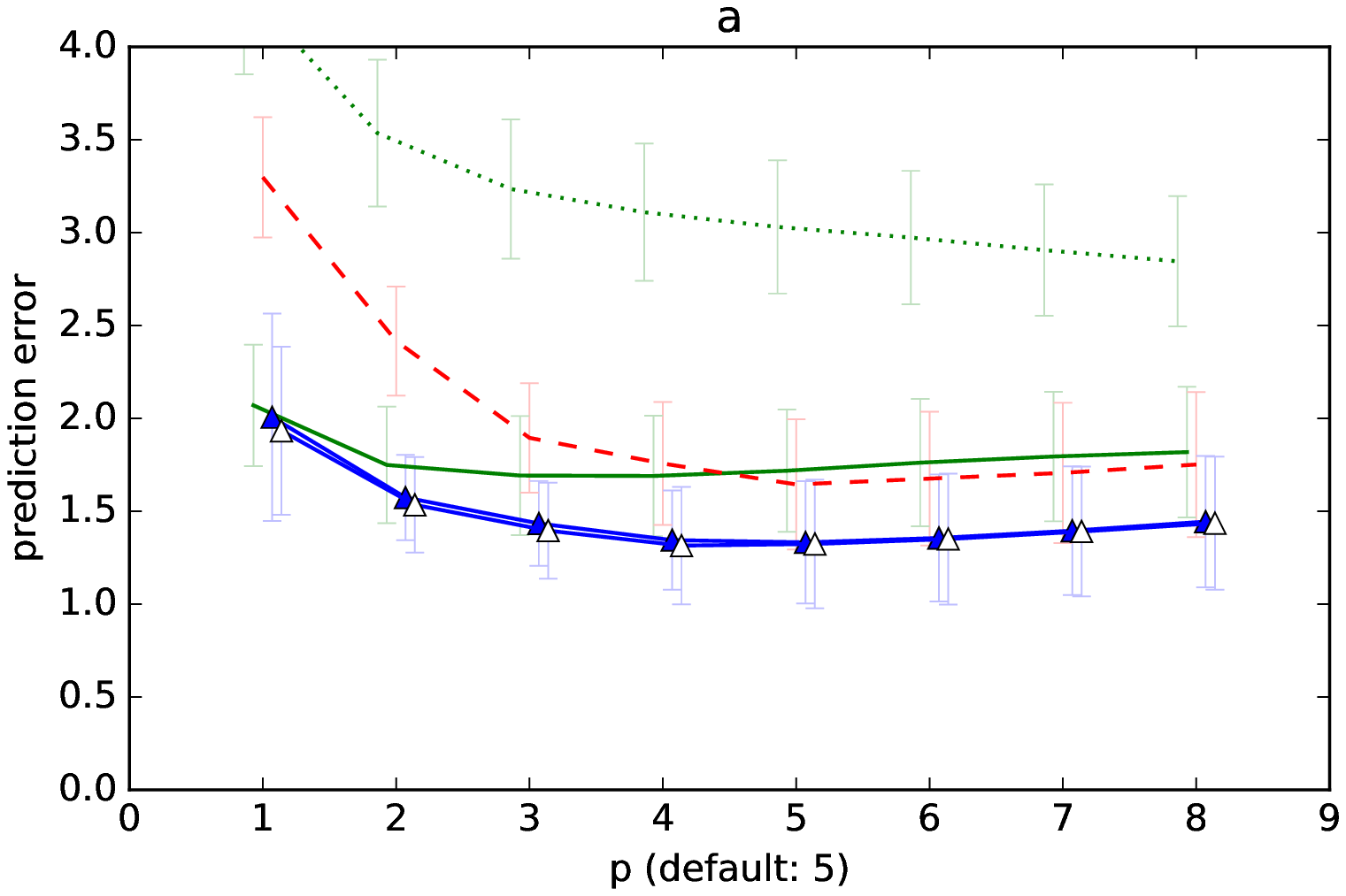}
\includegraphics[width=.45\textwidth]{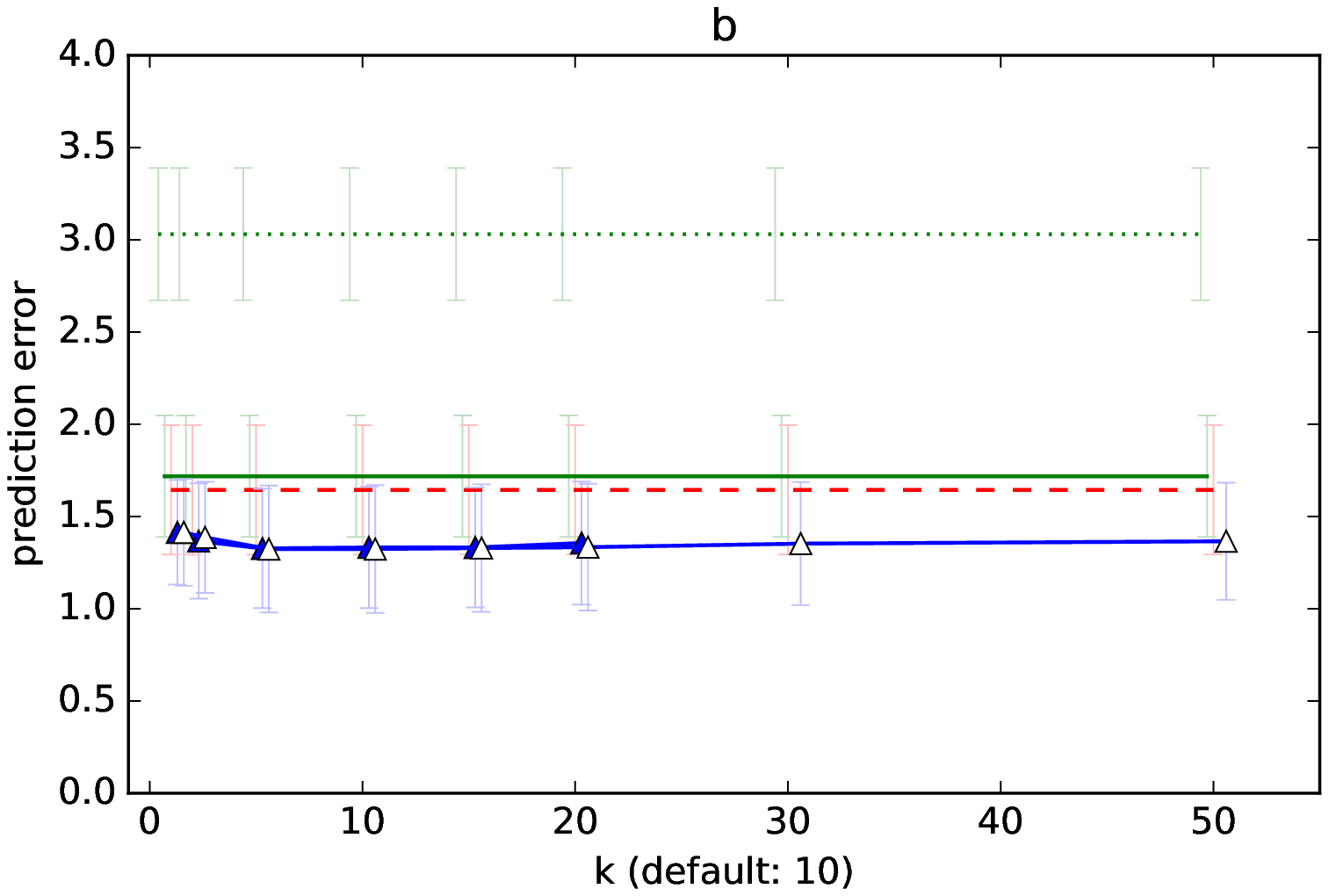}
\includegraphics[width=.45\textwidth]{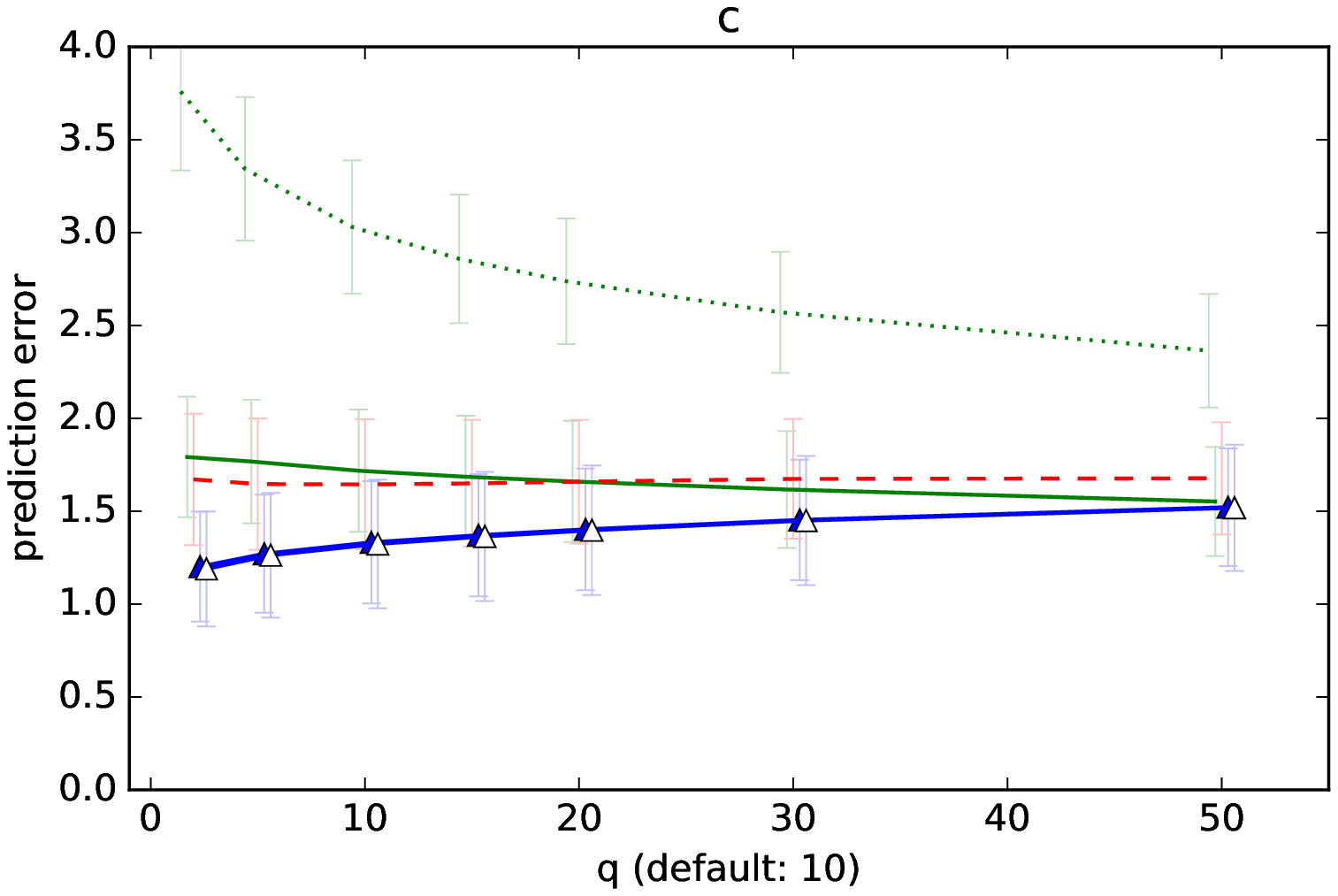}
\includegraphics[width=.45\textwidth]{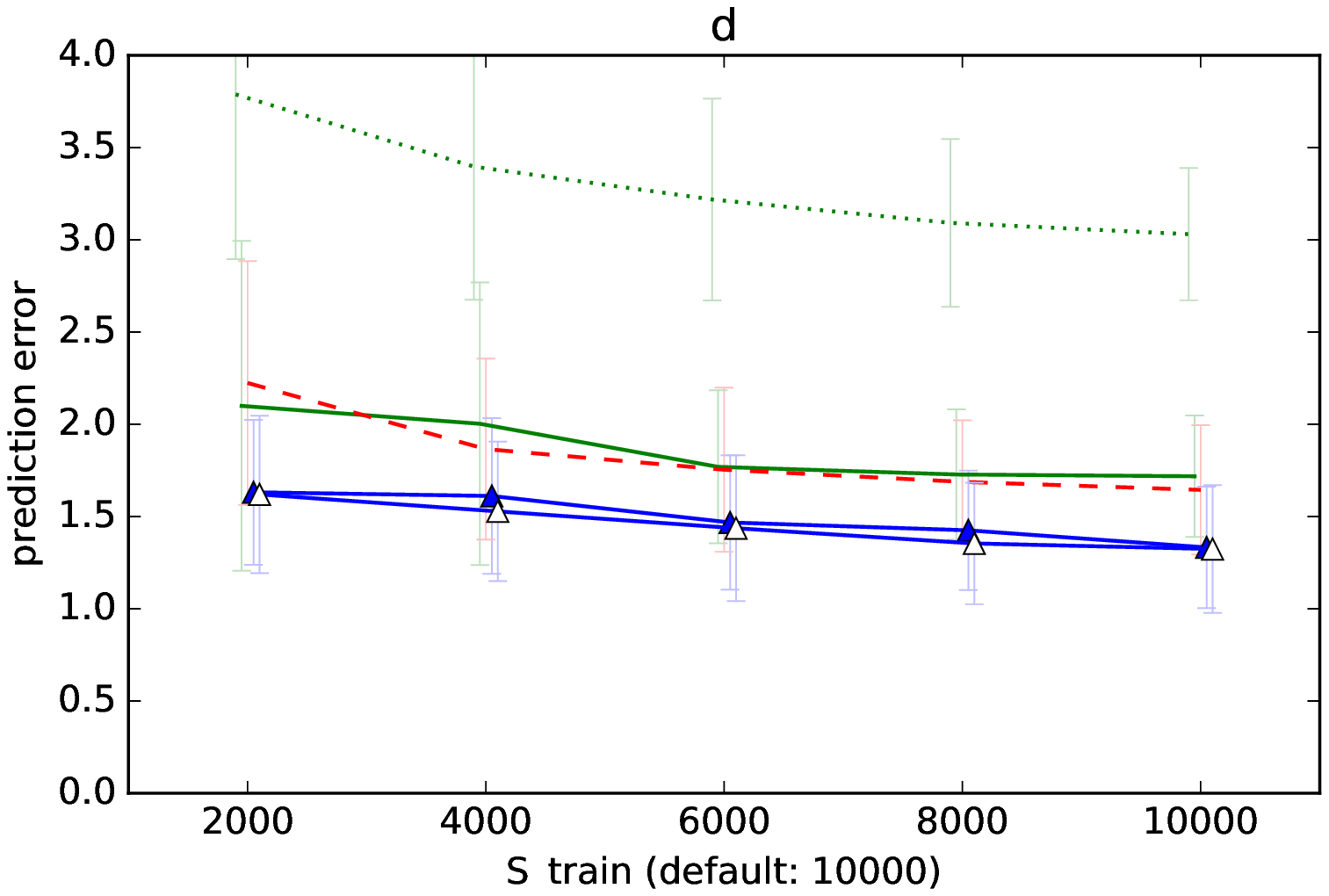}
\includegraphics[width=.45\textwidth]{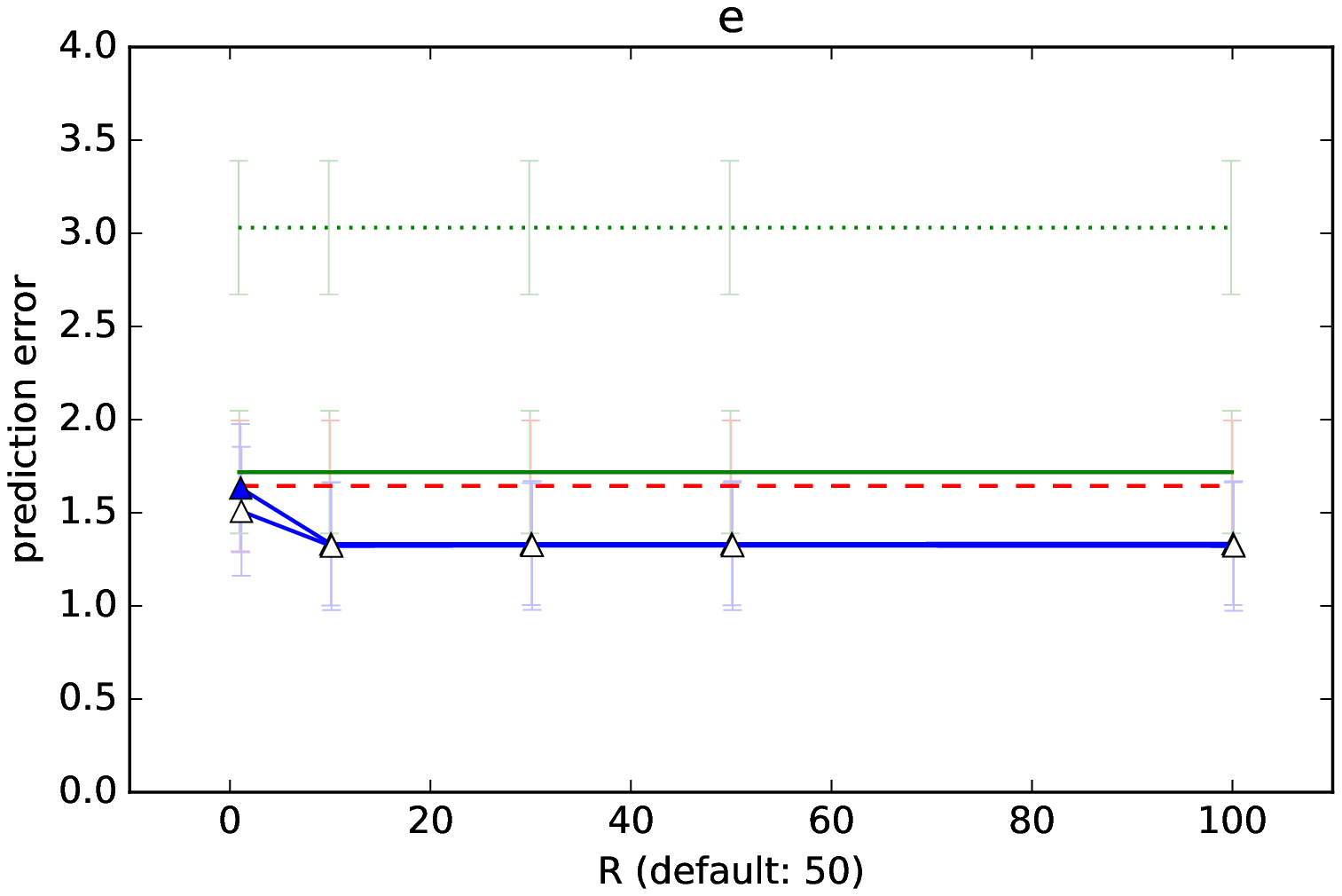}
\includegraphics[width=.45\textwidth]{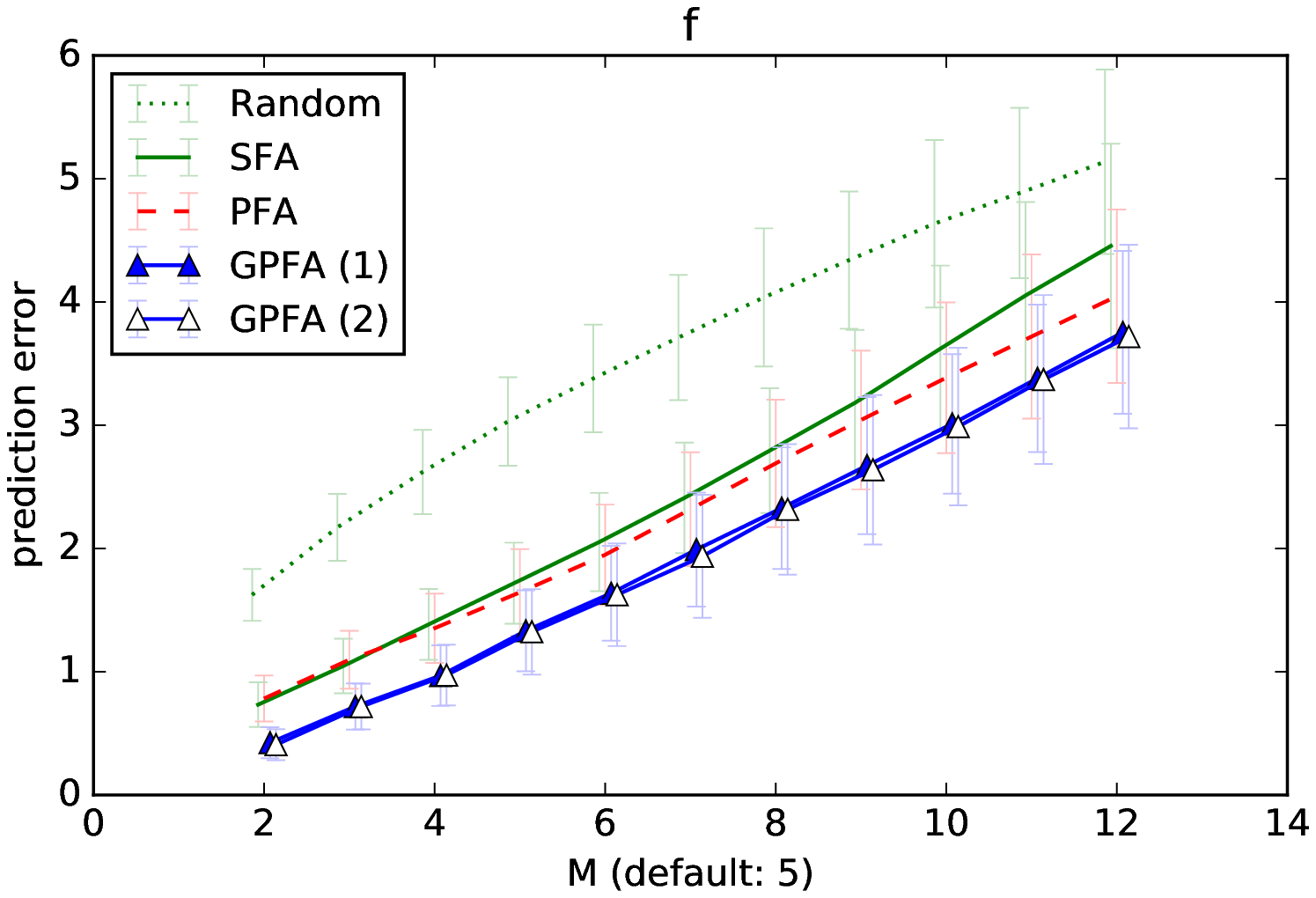}
\caption{Results for STFT \#1 (\emph{``piano''}): If not varied during the experiment, parameters were $p=5$, $k=10$, $q=10$, $S_{train}=10000$, $R=50$, $M=5$, and $K=10$. Slight x-shifts have been induced to separate error bars.}
\label{fig:results_1}
\end{figure}

\begin{figure}[htb]
\centering
\includegraphics[width=.45\textwidth]{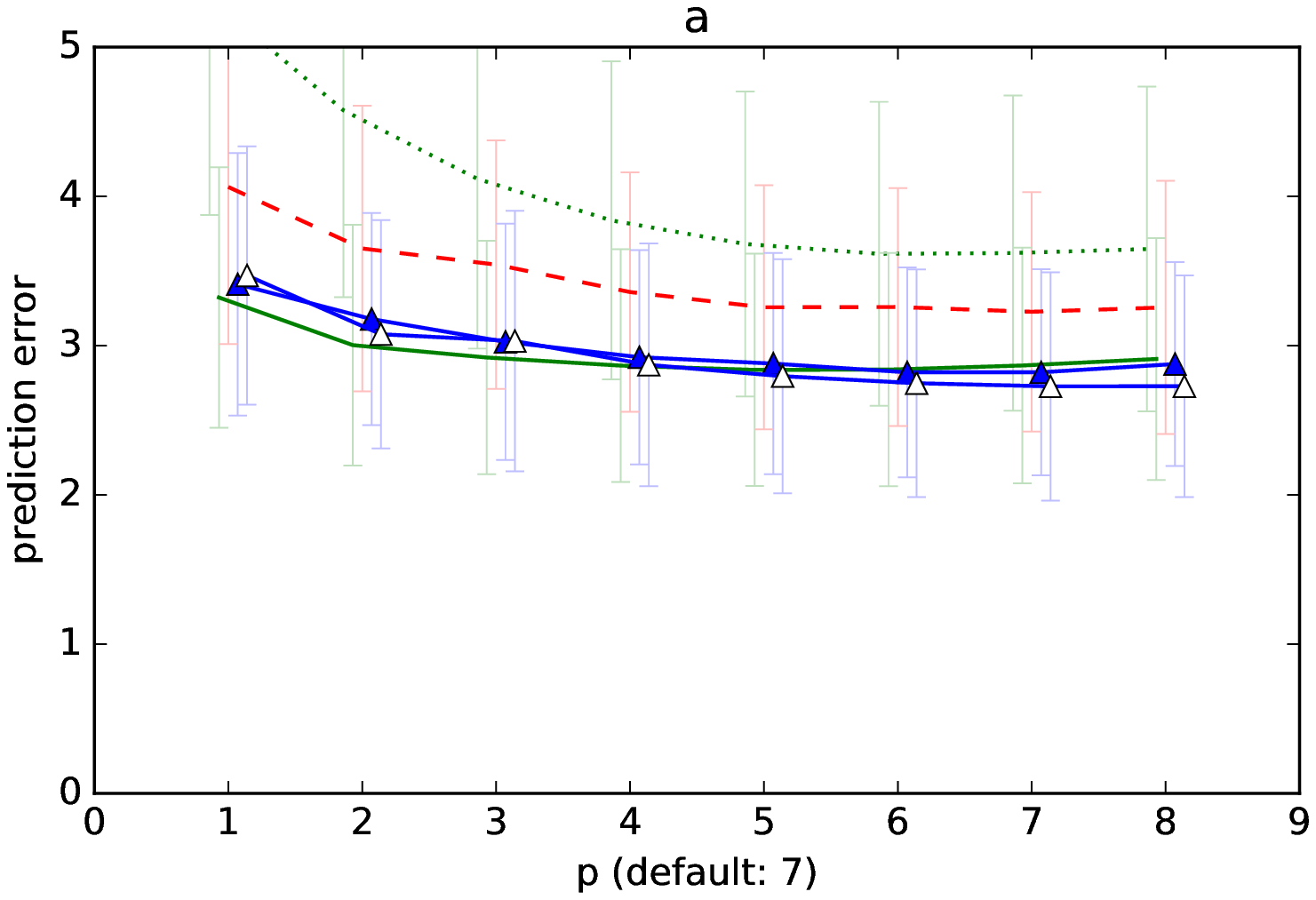}
\includegraphics[width=.45\textwidth]{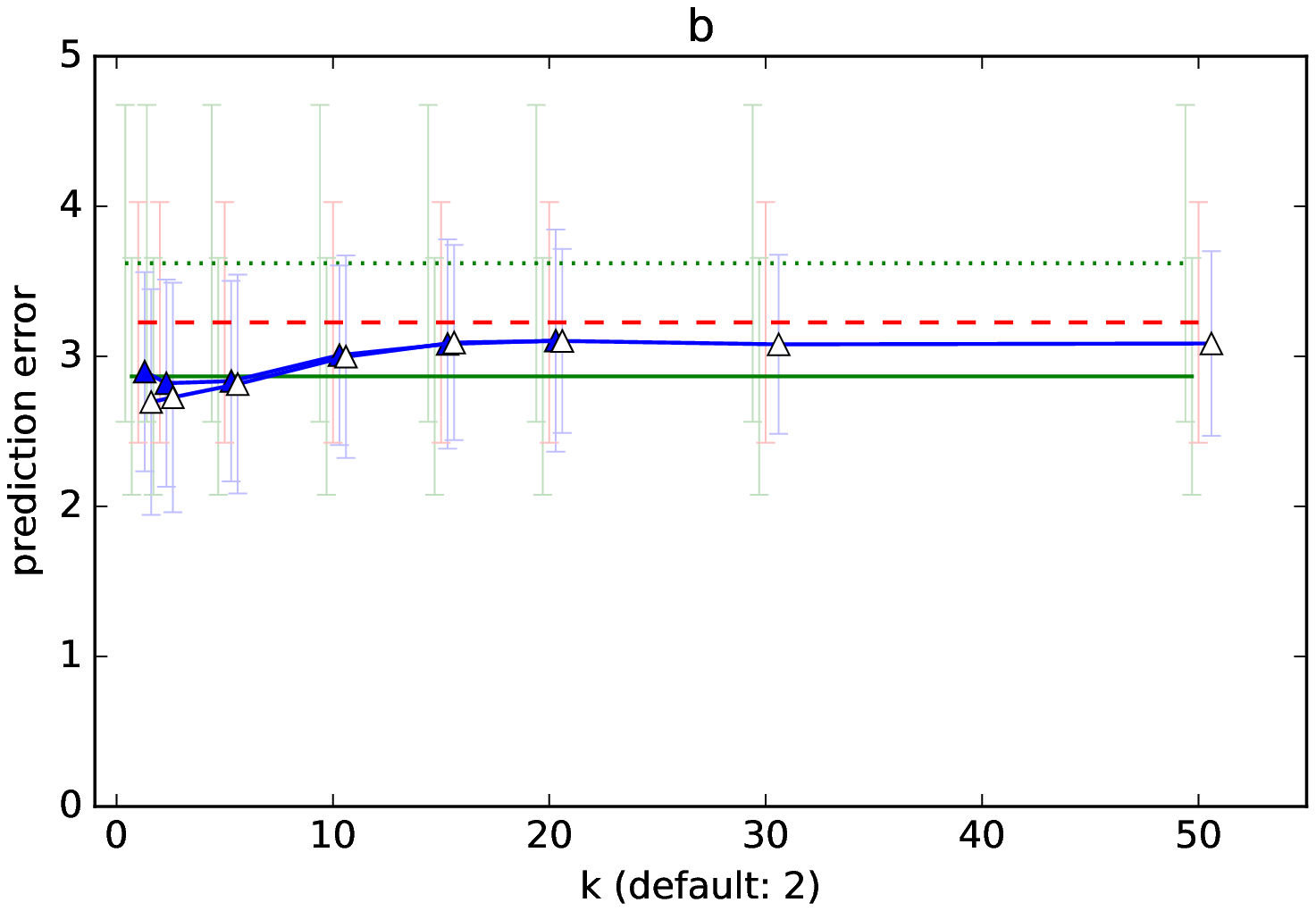}
\includegraphics[width=.45\textwidth]{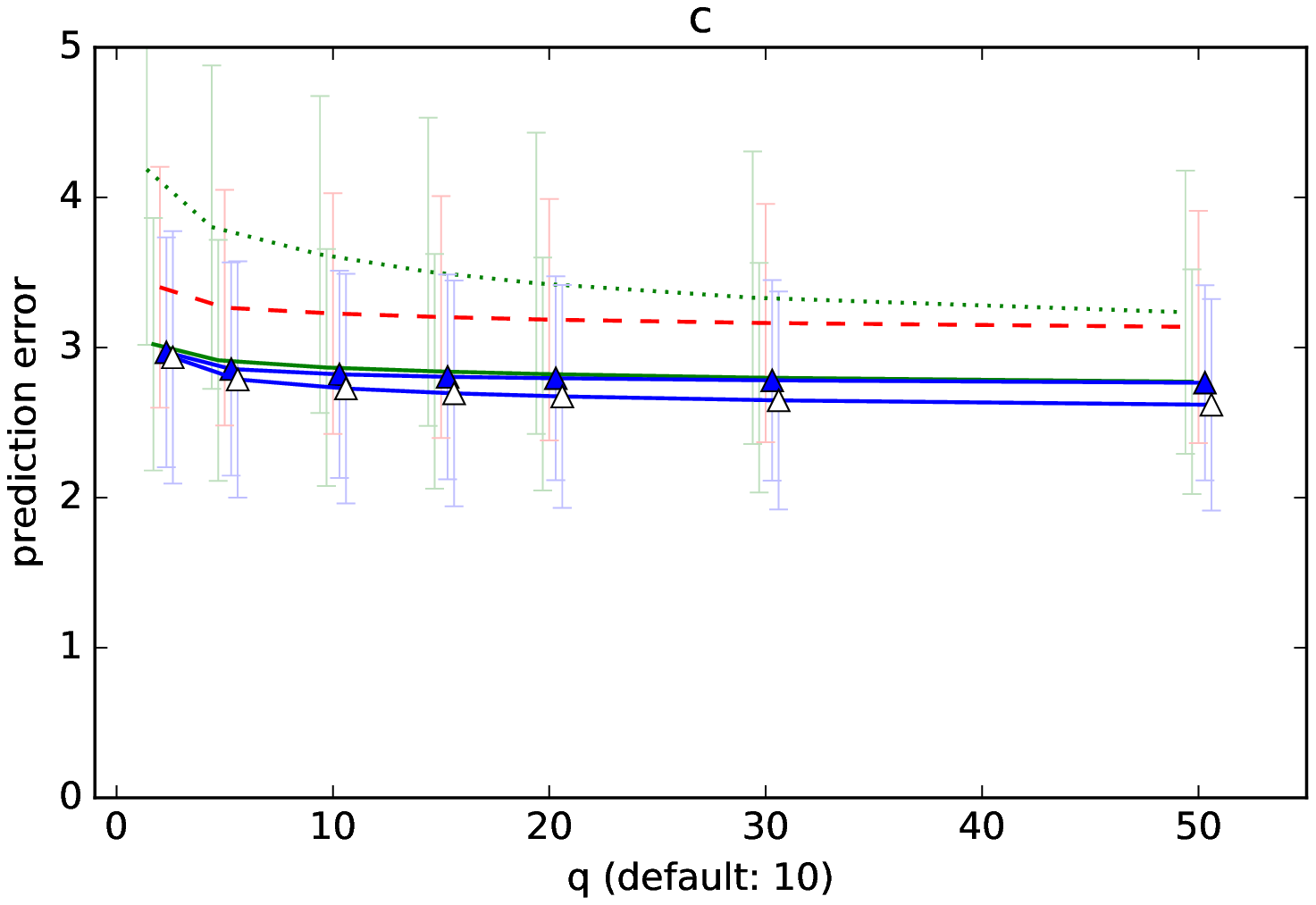}
\includegraphics[width=.45\textwidth]{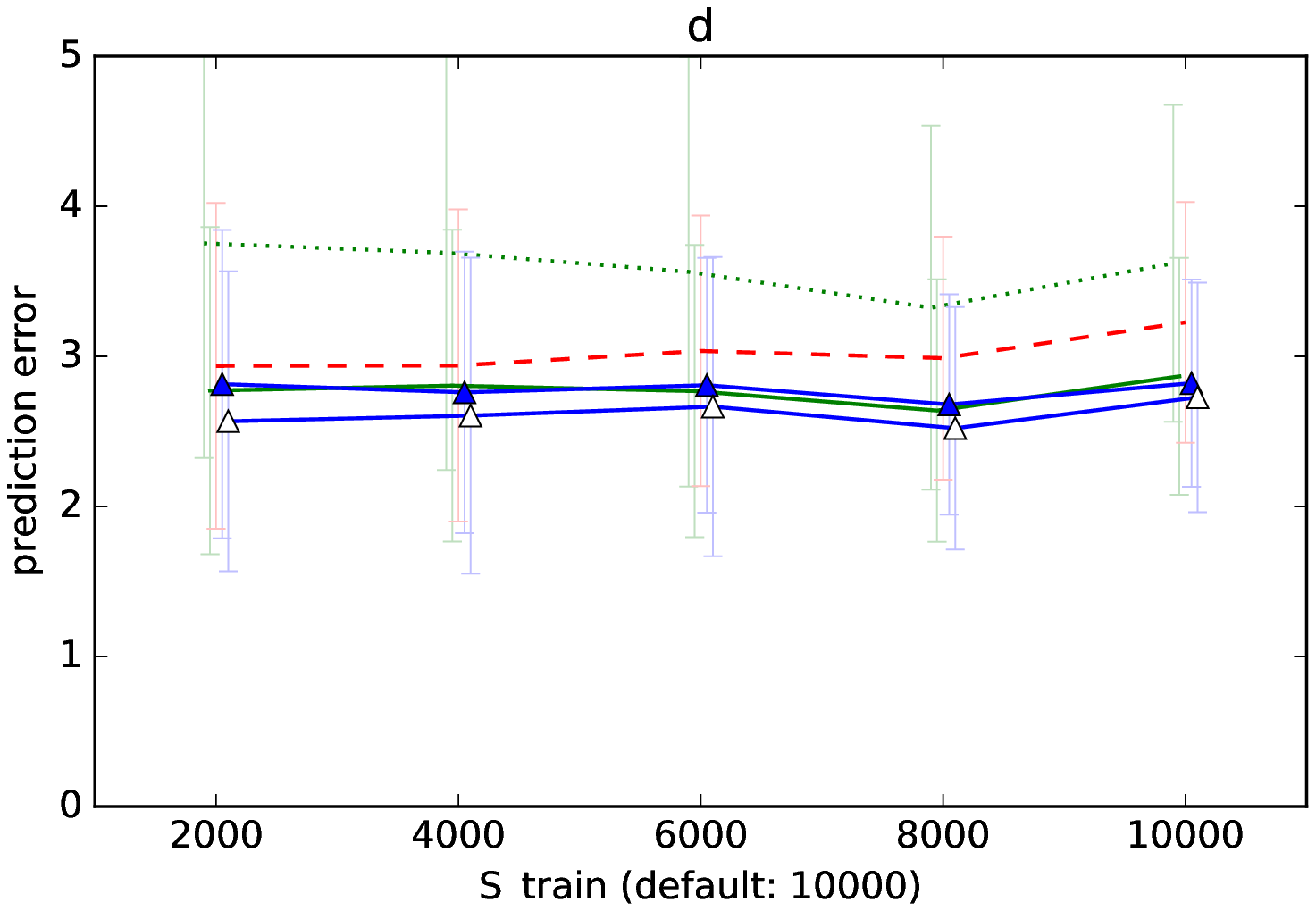}
\includegraphics[width=.45\textwidth]{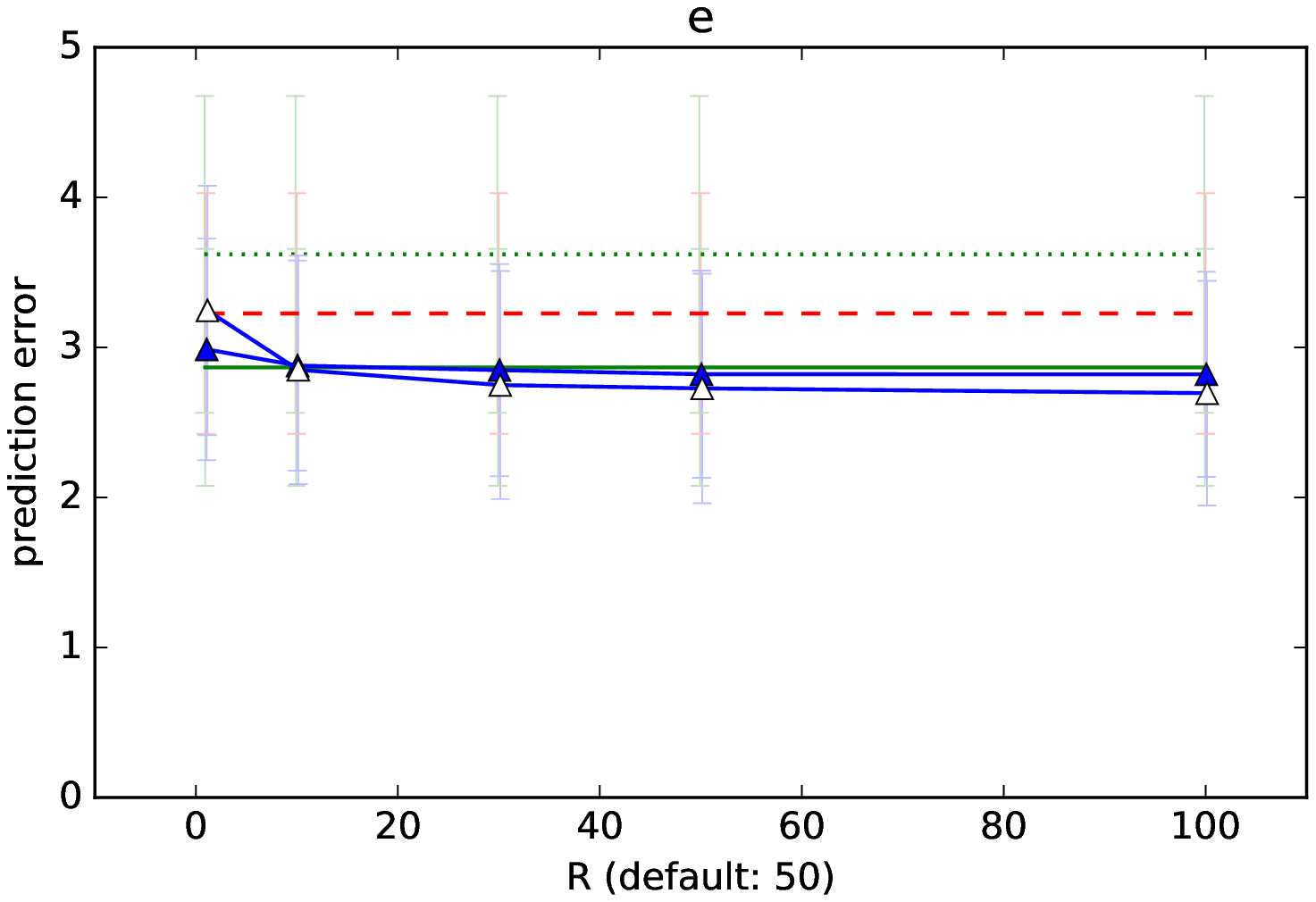}
\includegraphics[width=.45\textwidth]{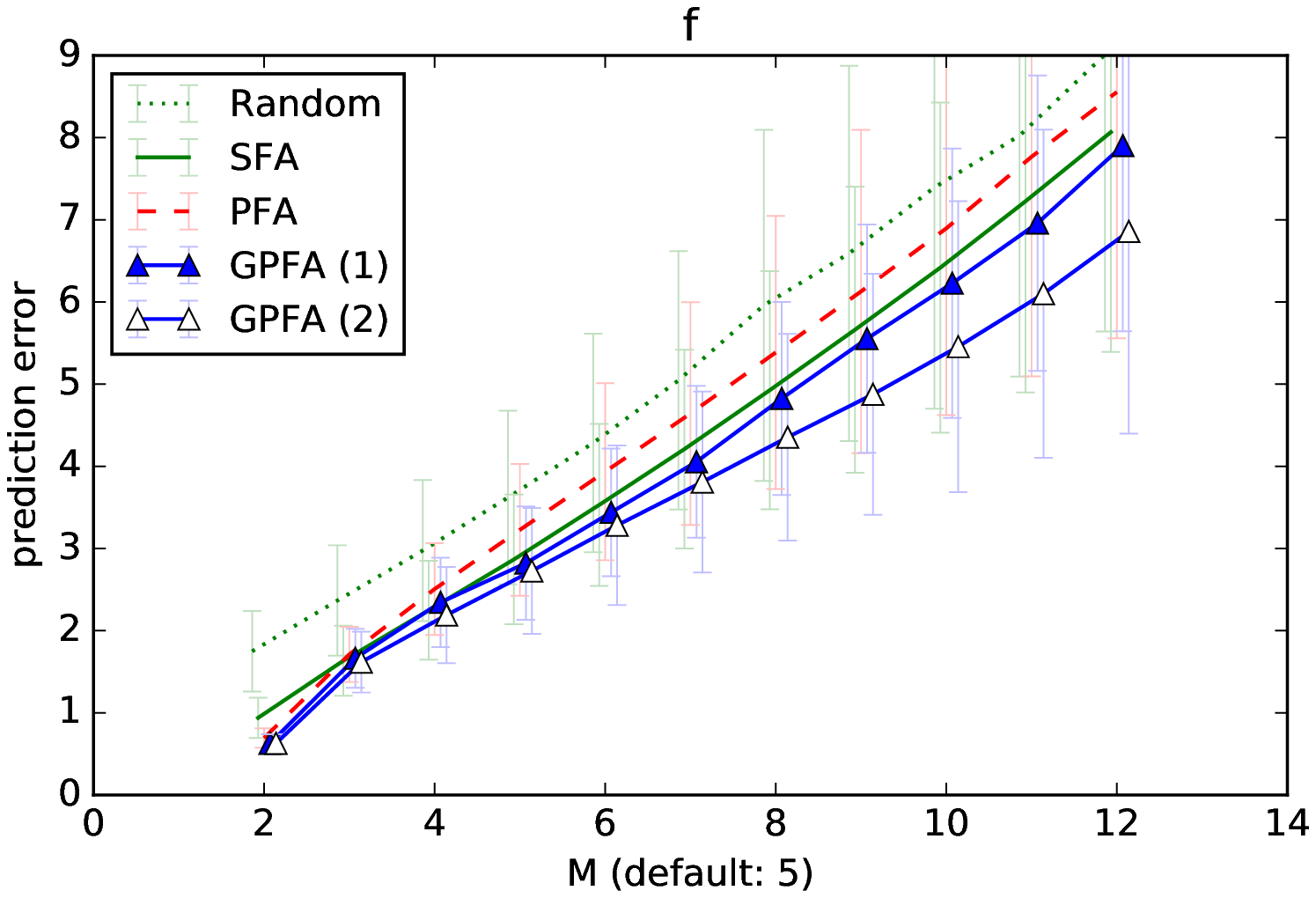}
\caption{Results for STFT \#2 (\emph{``bar''}): If not varied during the experiment, parameters were $p=7$, $k=2$, $q=10$, $S_{train}=10000$, $R=50$, $M=5$, and $K=10$. Slight x-shifts have been induced to separate error bars.}
\label{fig:results_2}
\end{figure}

\begin{figure}[htb]
\centering
\includegraphics[width=.45\textwidth]{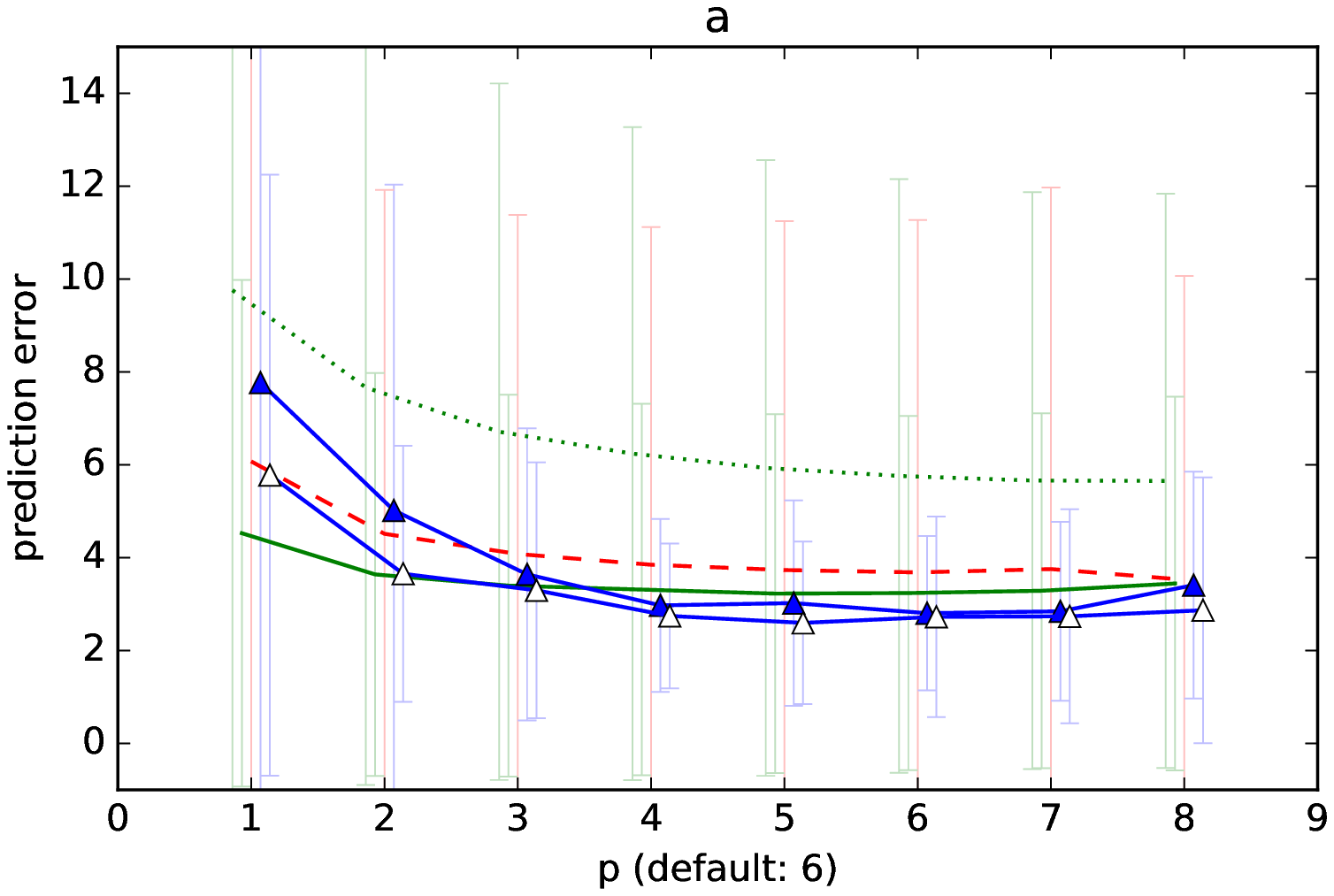}
\includegraphics[width=.45\textwidth]{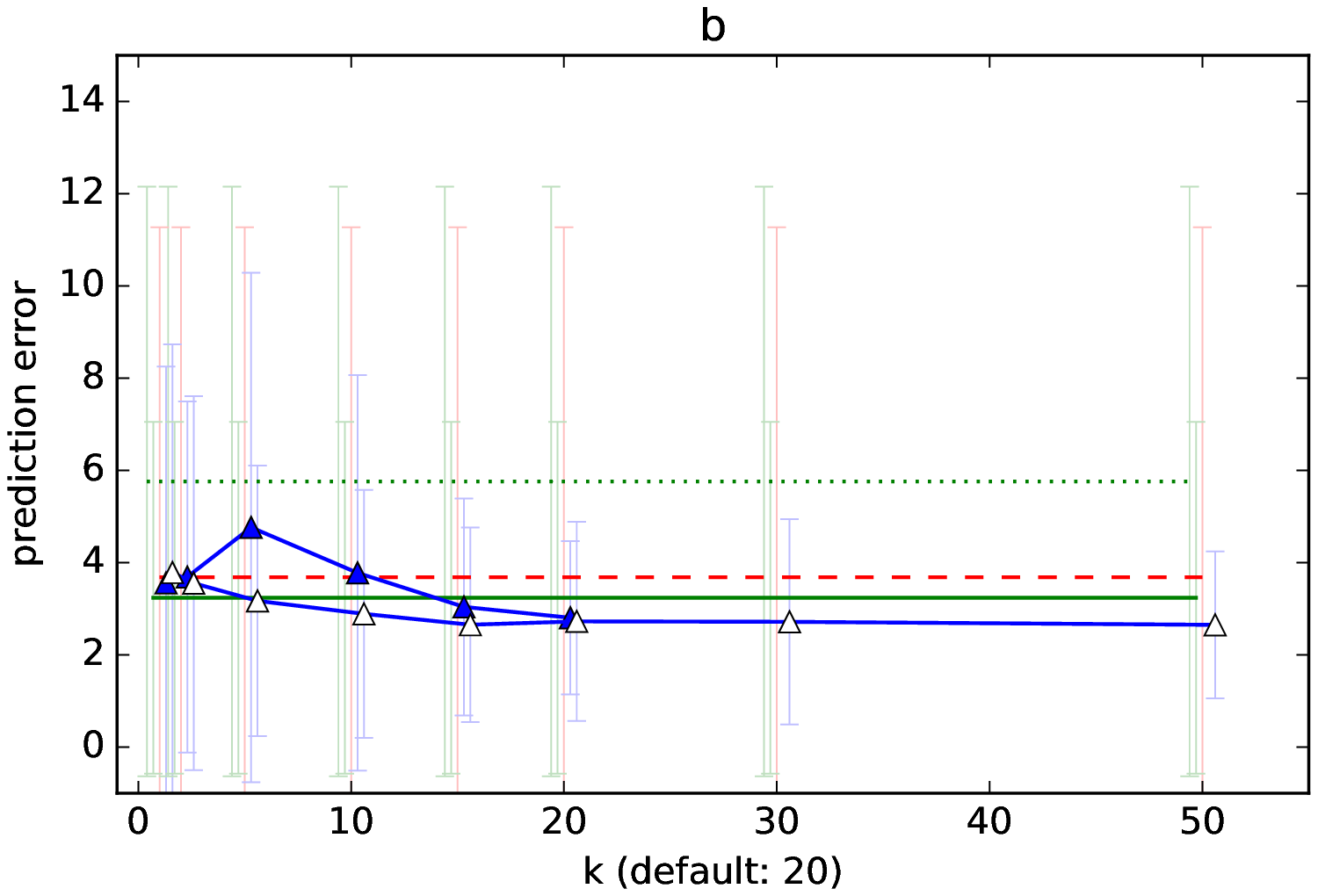}
\includegraphics[width=.45\textwidth]{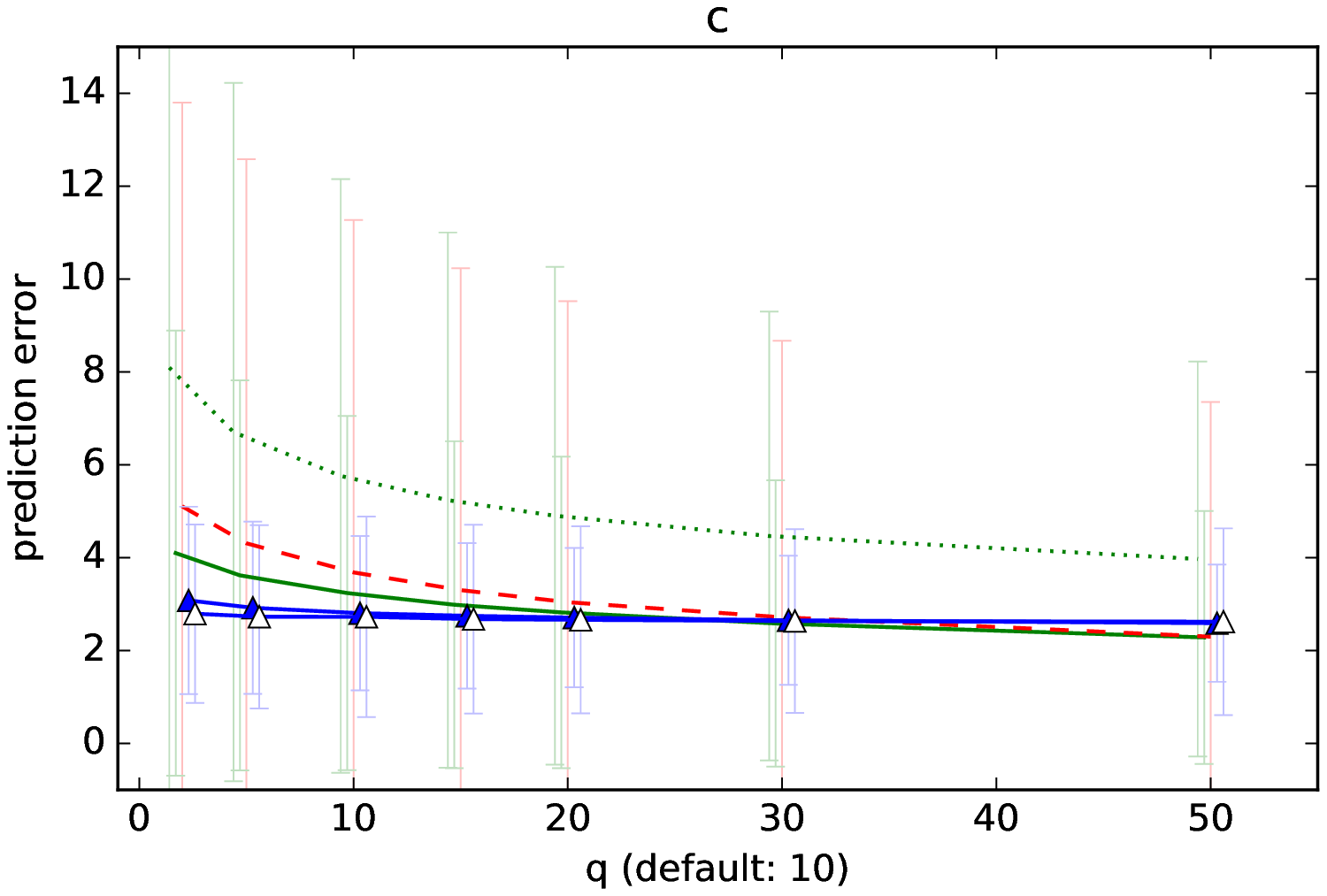}
\includegraphics[width=.45\textwidth]{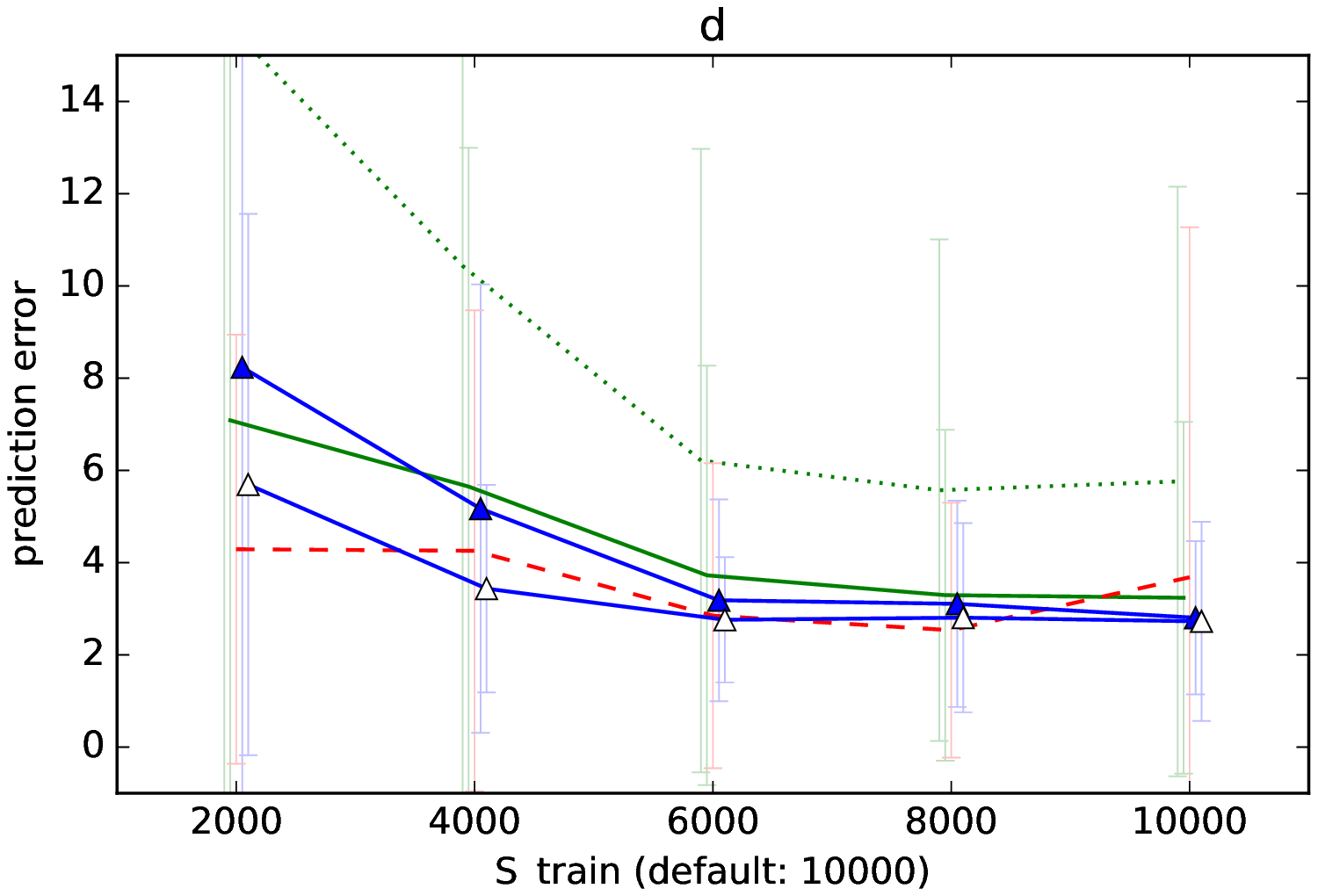}
\includegraphics[width=.45\textwidth]{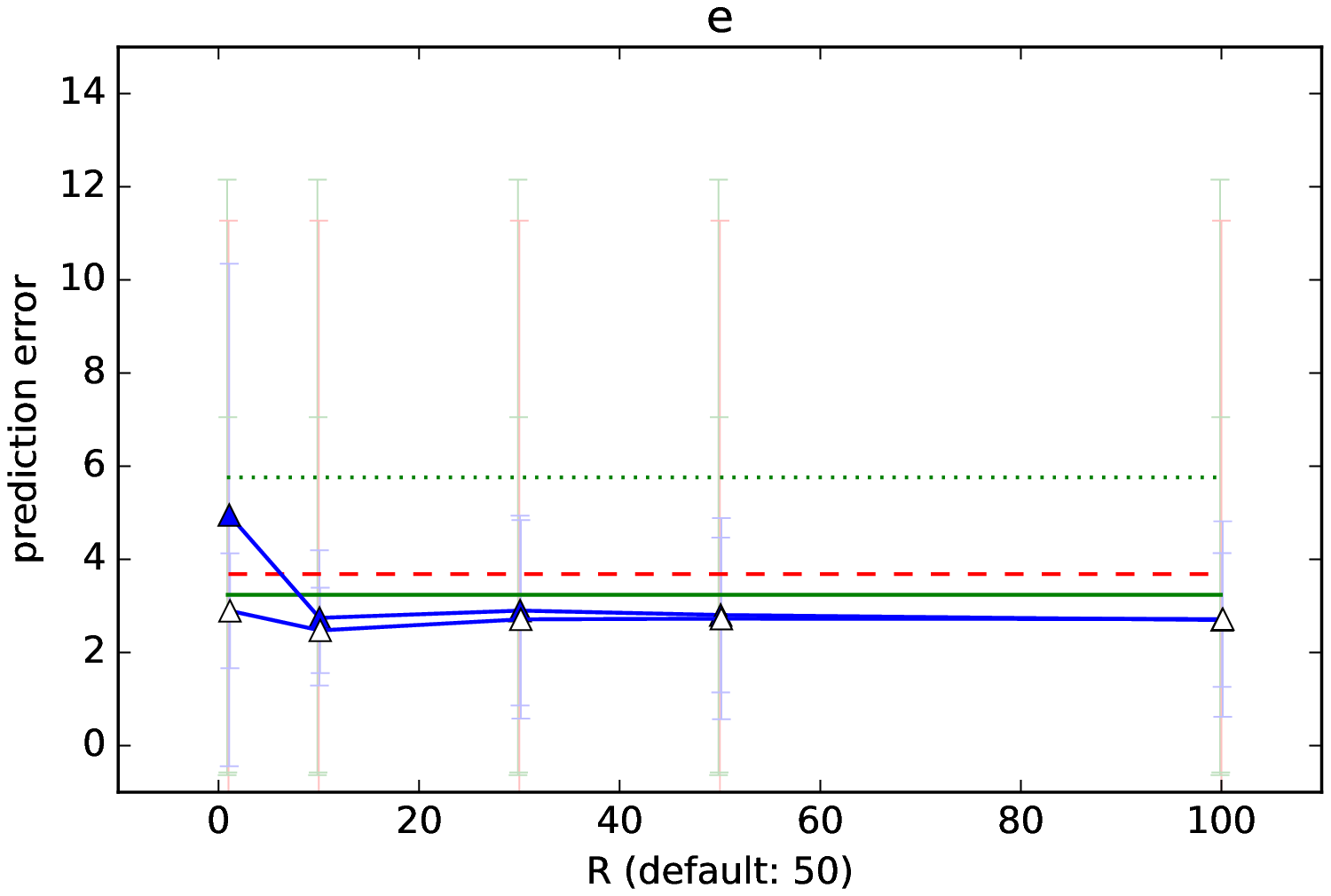}
\includegraphics[width=.45\textwidth]{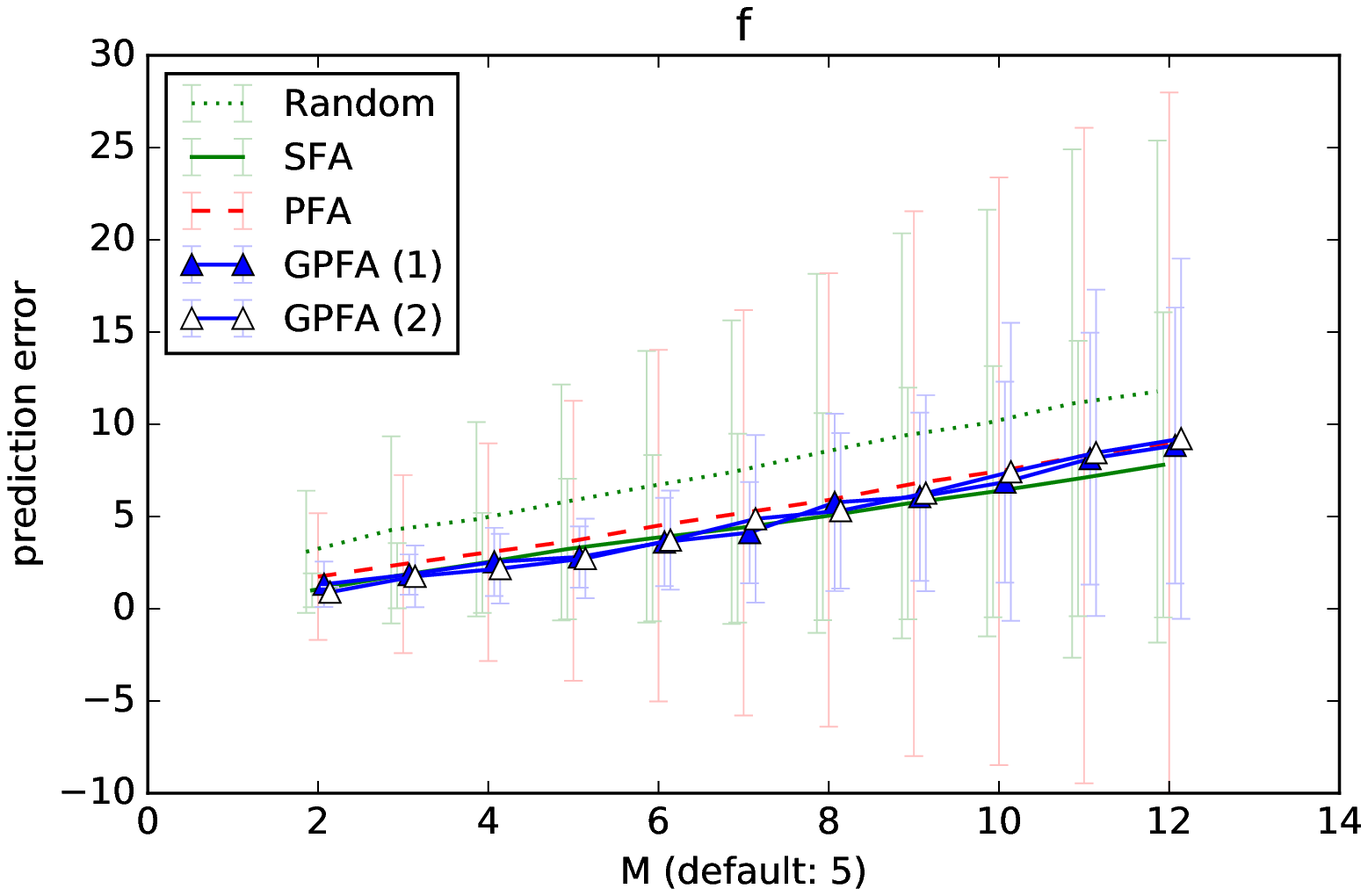}
\caption{Results for STFT \#3 (\emph{``forest''}): If not varied during the experiment, parameters were $p=6$, $k=20$, $q=10$, $S_{train}=10000$, $R=50$, $M=5$, and $K=10$. Slight x-shifts have been induced to separate error bars.}
\label{fig:results_3}
\end{figure}

The central results can be seen in Figures~\ref{fig:results_1}-\ref{fig:results_3}f in terms of the predictability of the components extracted by the different algorithms in dependence of their dimensionality $M$. The other plots show how the results change with the individual parameters. Increasing the number of past time steps $p$ tends to improve the results first but may let them decrease later (see Figures~\ref{fig:results_1}-\ref{fig:results_3}a). Presumably, because higher numbers of $p$ make the models more prone to overfitting. The neighborhood size $k$ had to be selected carefully for each of the different datasets. While its choice was not critical on the first dataset, the second dataset benefited from low values for $k$ and the third one from higher values (see Figures~\ref{fig:results_1}-\ref{fig:results_3}b). Similar, the neighborhood size $q$ for calculating the final predictability of the results had different effects for different datasets (see Figures~\ref{fig:results_1}-\ref{fig:results_3}c). At this point it's  difficult to favor one value over another, which is why we kept $q$ fixed to $q=10$. As expected, results tend to improve with increasing numbers of training samples $S_{train}$ (see Figures~\ref{fig:results_1}-\ref{fig:results_3}d). Similarly, results first improve with the number of iterations $R$ for GPFA and then remain stable (see Figures~\ref{fig:results_1}-\ref{fig:results_3}e). We take this as evidence for the viability of the iteration heuristic motivated in Section~\ref{sec:iterated_gpfa}.

To gauge the statistical reliability of the results, we applied the Wilcoxon signed-rank test, testing the null hypothesis that the results for different pairs of algorithms actually come from the same distribution. We tested this hypothesis for each data set for the experiment with default parameters, i.e., for the results shown in Figures~\ref{fig:results_1}-\ref{fig:results_3}f with $M=5$. As can be seen from the $p$-values in Table~\ref{tbl:wilcoxon_stft}, the null hypothesis can be rejected with certainty in many cases, which confirms that GPFA~(2) learned the most predictable features on two of three datasets. For GPFA~(1) the results are clear for the first dataset as well for the second in comparison to PFA. It remains a small probability, however, that the advantage compared to SFA on the second dataset is only due to chance. For the large third dataset, all algorithms produce relatively similar results with high variance between experiments. It depends on the exact value of $M$ if SFA or GPFA produced the best results. For $M=5$ GPFA happened to find slightly more predictable results (not highly significant though as can be seen in Table~\ref{tbl:wilcoxon_stft}). But in general we don't see a clear advantage of GPFA on the third dataset.

\begin{table}[htb]
\centering
\caption{$p$-values for the Wilcoxon signed-rank test which tests the null hypothesis that a pair of samples come from the same distribution. Values refer to the experiments shown in Figures~\ref{fig:results_1}-\ref{fig:results_3}f with $M=5$. Row and column indicate the pair of algorithms compared. $p$-values that show a significant ($p \le 0.01$) advantage of GPFA over the compared algorithm are printed \textbf{bold}.}
\label{tbl:wilcoxon_stft}
\begin{tabular}{l|ll|ll|ll}
 & \multicolumn{2}{|c|}{STFT \#1} & \multicolumn{2}{|c|}{STFT \#2} & \multicolumn{2}{|c}{STFT \#3} \\
 &  SFA & PFA & SFA & PFA & SFA & PFA \\
\hline
GPFA~(1) & \textbf{0.00} & \textbf{0.00} & 0.18 & \textbf{0.00} & 0.43 & 0.09 \\
GPFA~(2) & \textbf{0.00} & \textbf{0.00} & \textbf{0.00} & \textbf{0.00} & 0.38 & 0.17      
\end{tabular}
\end{table}

\subsection{Visual data}

\begin{figure}[htbp]
\centering
\includegraphics[width=.5\textwidth]{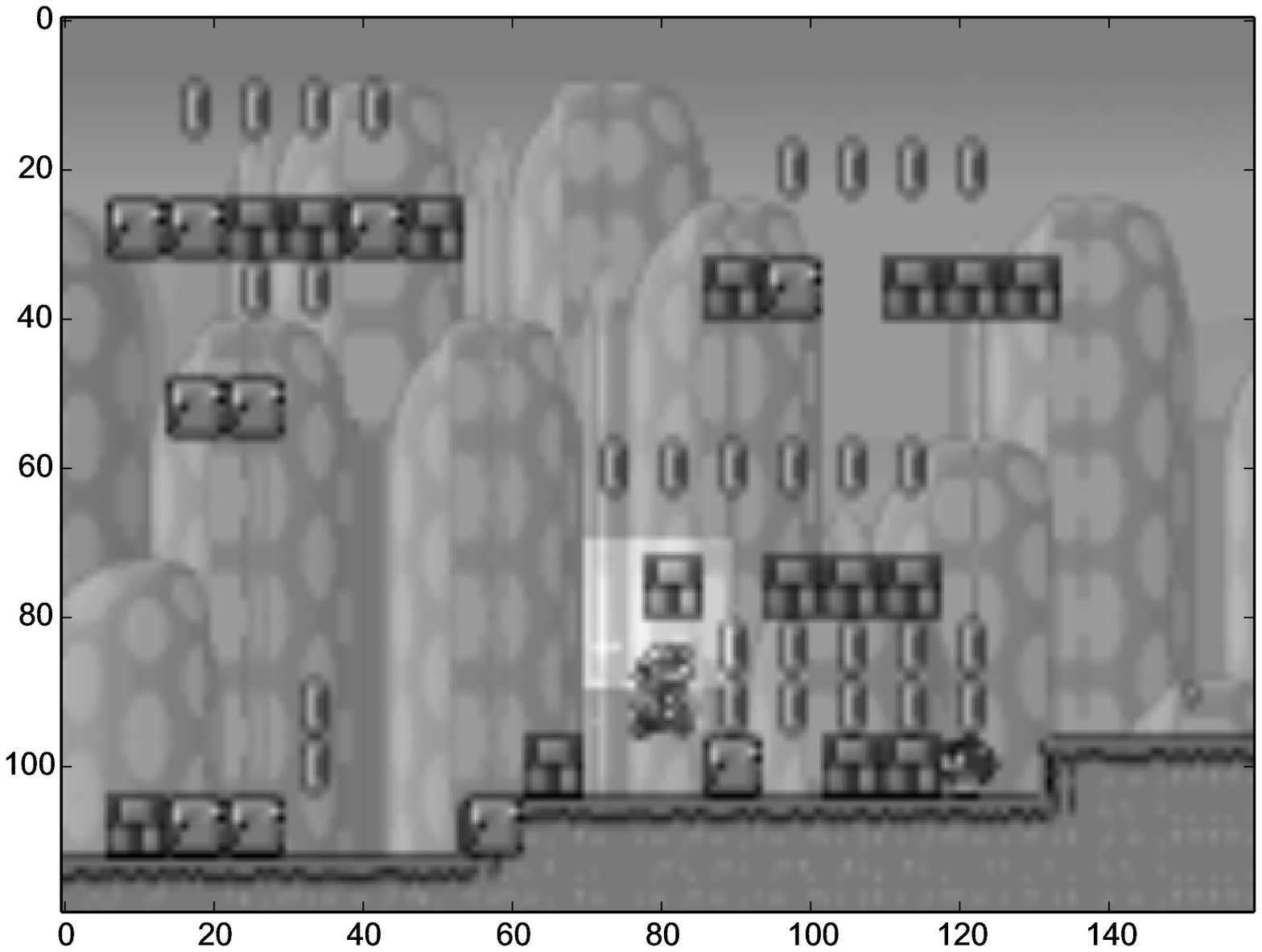}
\caption{An example frame generated by a modified simulator from the Mario AI challenge. The highlighted square indicates the $400$ pixels extracted for the experiment.}
\label{fig:example_mario}
\end{figure}

\begin{figure}[htb]
\centering
\includegraphics[width=.45\textwidth]{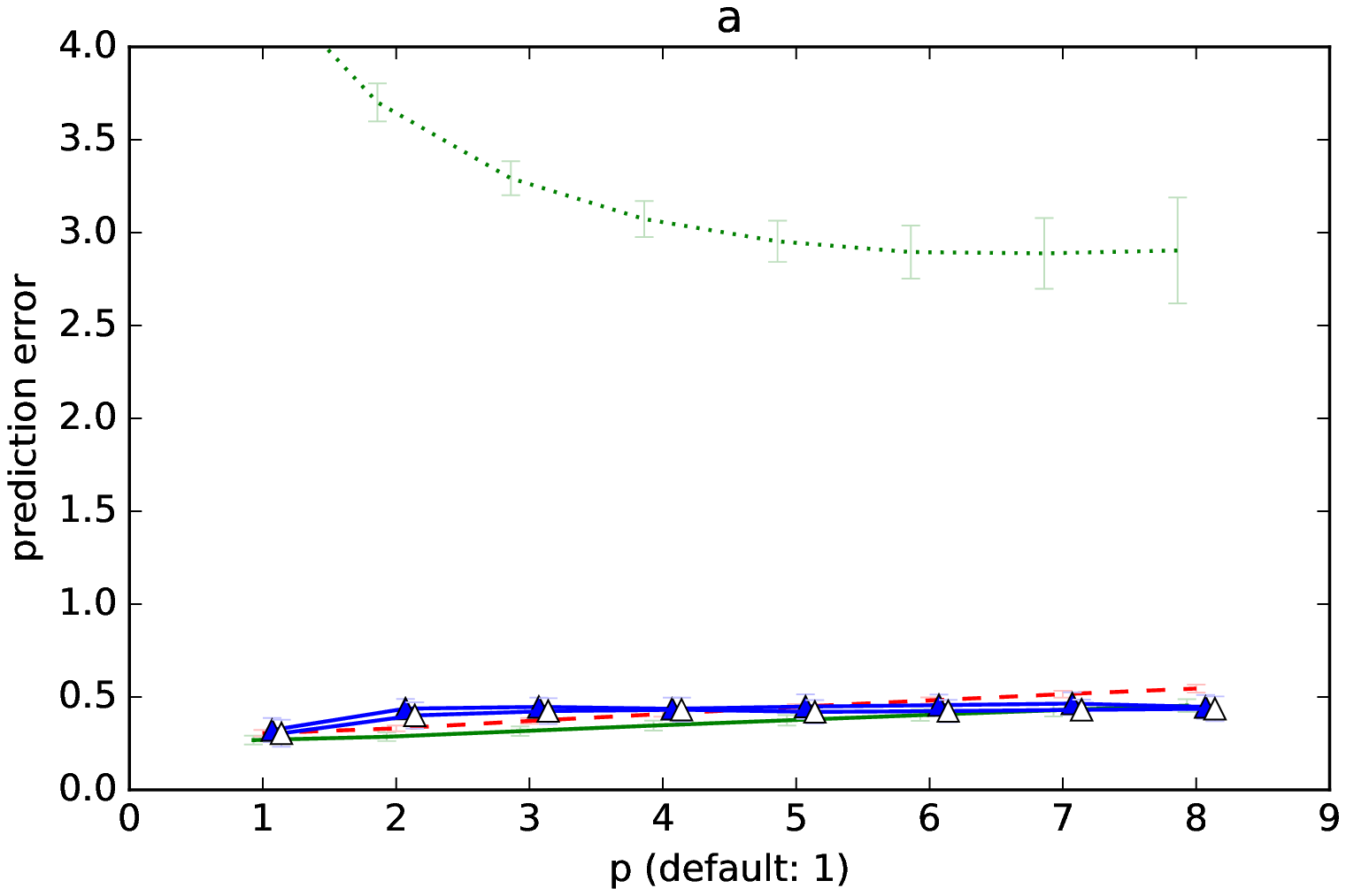}
\includegraphics[width=.45\textwidth]{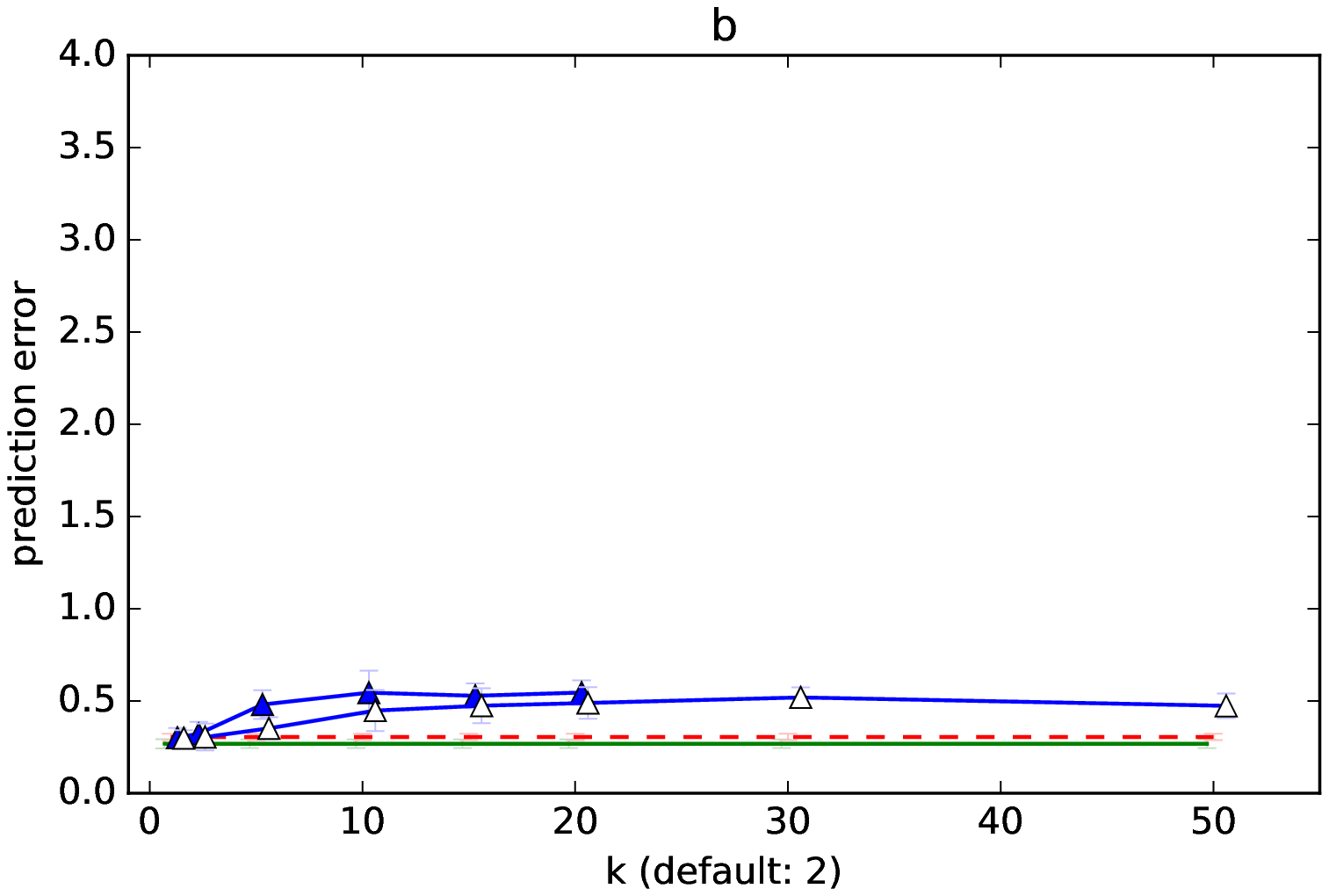}
\includegraphics[width=.45\textwidth]{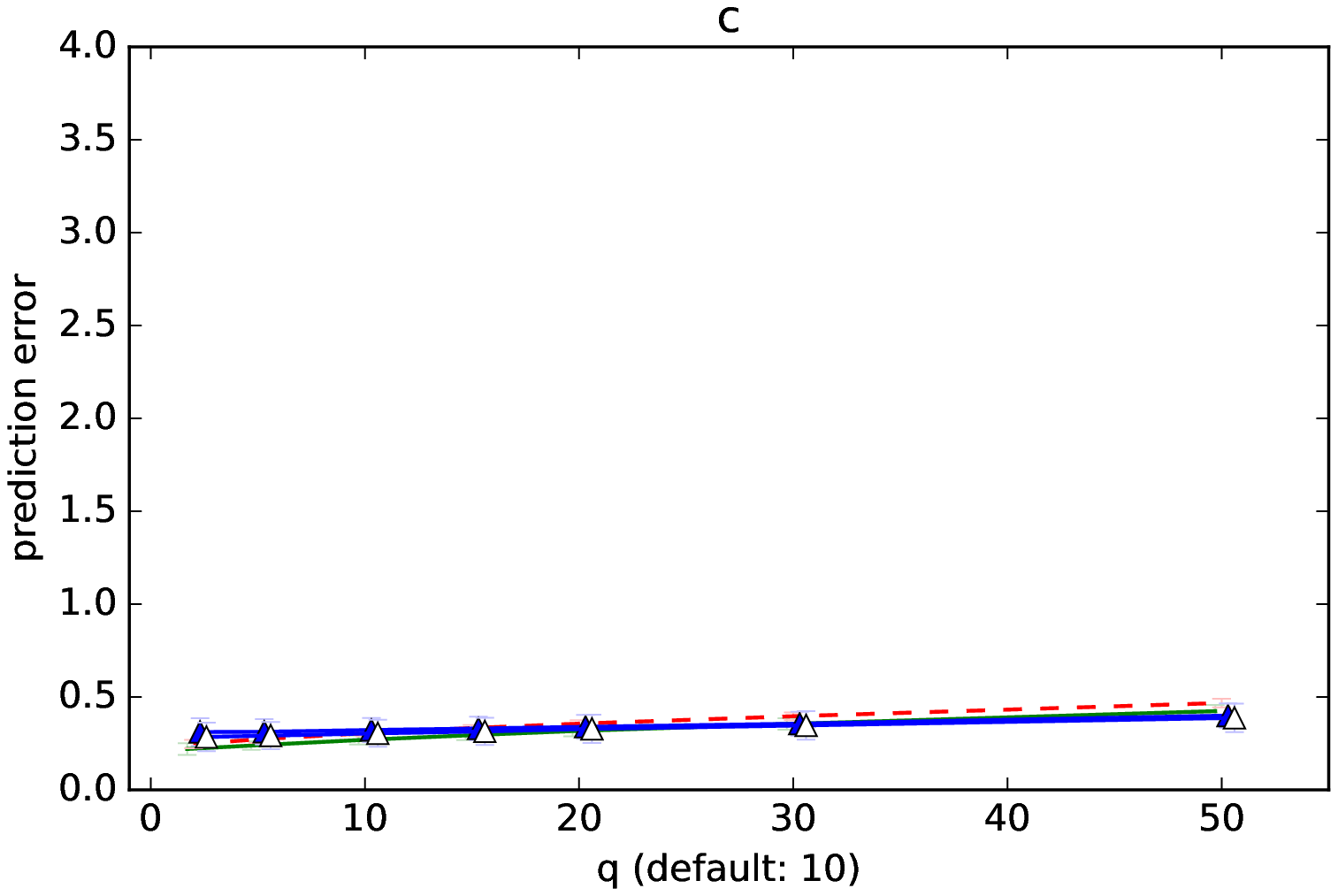}
\includegraphics[width=.45\textwidth]{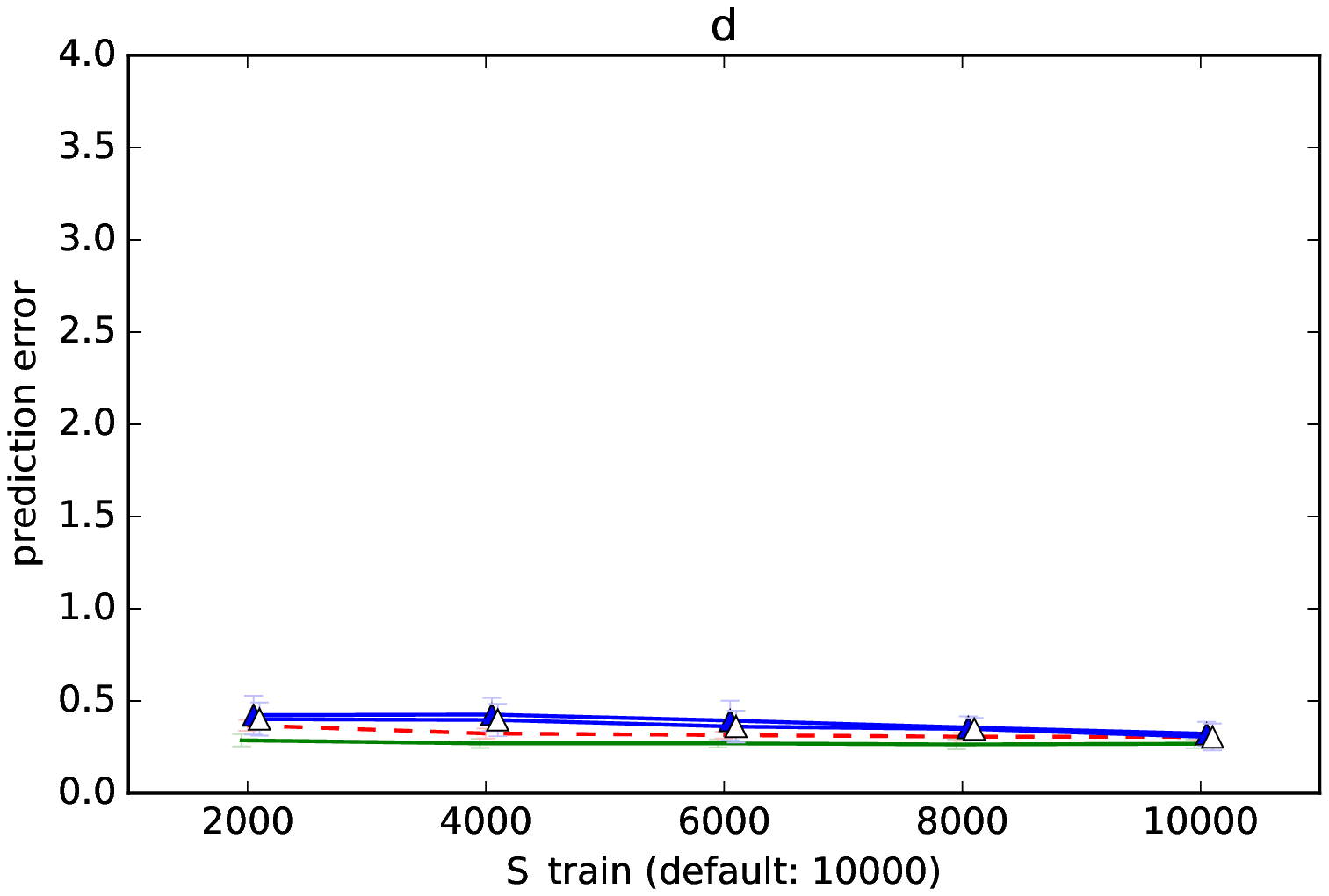}
\includegraphics[width=.45\textwidth]{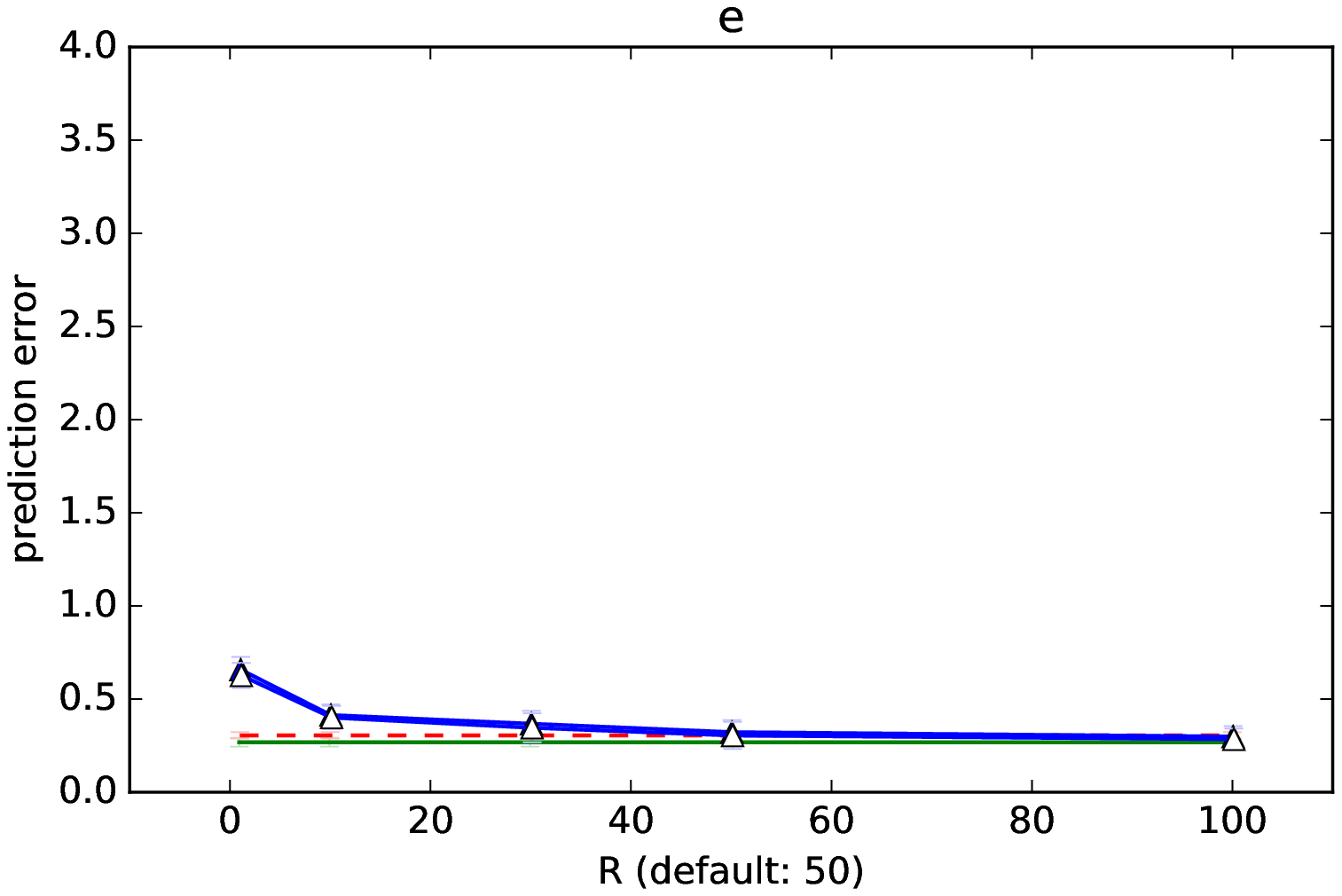}
\includegraphics[width=.45\textwidth]{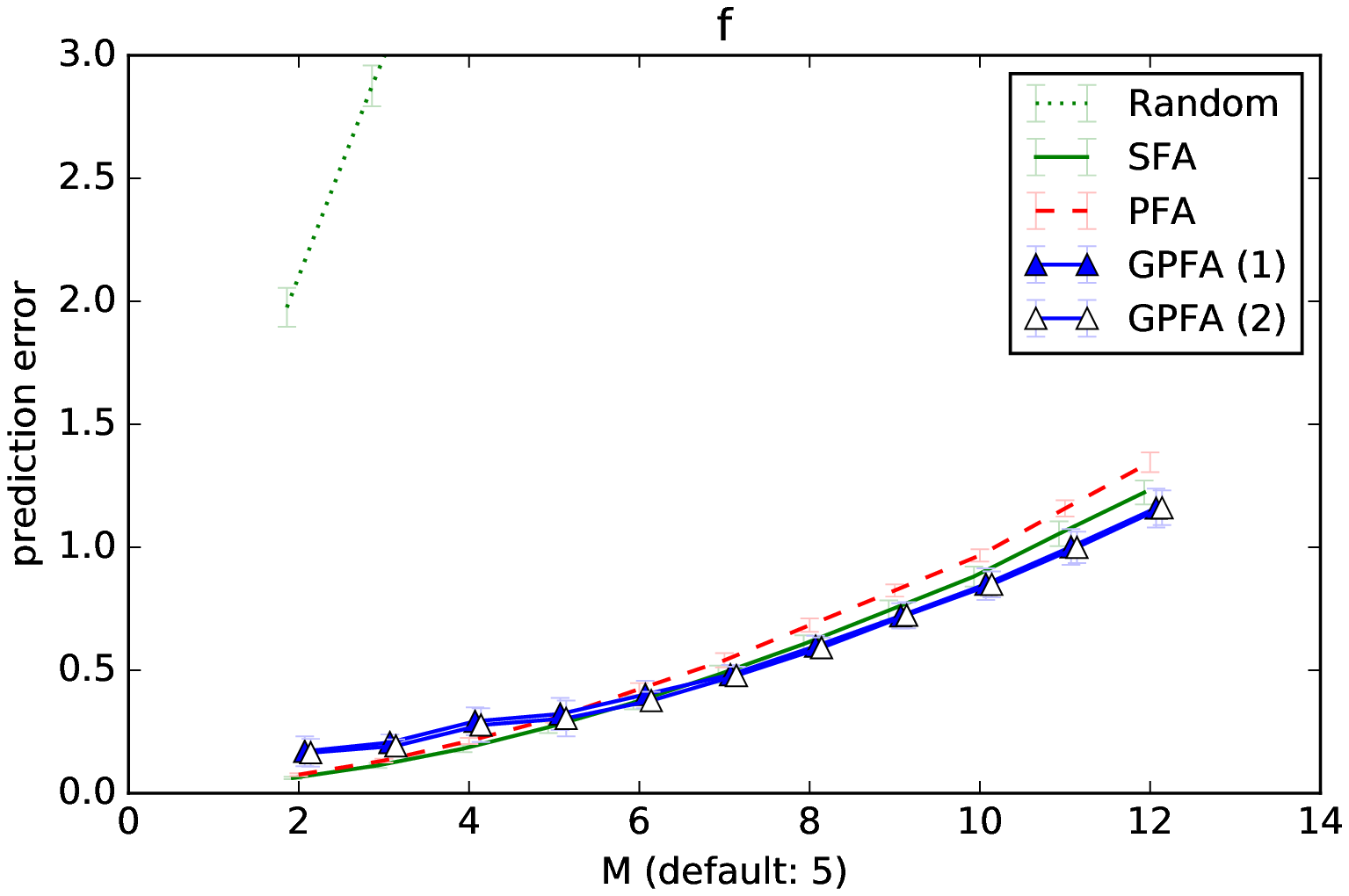}
\caption{Results for visual dataset (\emph{``Super Mario''}): If not varied during the experiment, parameters were $p=1$, $k=2$, $q=10$, $S_{train}=10000$, $R=50$, $M=5$, and $K=1$. Slight x-shifts have been induced to separate error bars.}
\label{fig:results_4}
\end{figure}

A third experiment was conducted on a visual dataset. We modified the simulator from the \emph{Mario AI challenge}~\citep{KarakovskiyTogelius-2012} to return raw visual input in gray-scale without text labels. The raw input was scaled from $320 \times 240$ down to $160 \times 120$ dimensions and then the final data points were taken from a small window of $20 \times 20 = 400$ pixels at a position where much of the game dynamics happened (see Figure~\ref{fig:example_mario} for an example). As with the auditory datasets, for each experiment $S_{train} = 10000$ successive training and $S_{test} = 5000$ non-overlapping test frames were selected randomly and PCA was applied to both, preserving $99\%$ of the variance. Eventually, $M$ predictable components were extracted by each of the algorithms and evaluated with respect to their predictability~\eqref{eq:knn_estimate}. Parameters $p$ and $k$ again were selected from a range of candidate values to yield the best results (see Figures~\ref{fig:results_4}a-b).

Two things are apparent from the results as shown in Figure~\ref{fig:results_4}. First, the choice of parameters was less critical compared to the auditory datasets. And second, all compared algorithms show quite similar results in terms of their predictability. GPFA only is able to find features slightly more predictable than those of SFA for higher values of $M$ (see Figure~\ref{fig:results_4}f). Again, this observation is highly significant with a Wilcoxon $p$-value of $0.00$ for $M=12$.

\section{Discussion and Future work}
\label{sec:discussion}

In the previous section we saw that GPFA produced the most predictable features on a toy example with a certain kind of predictable noise as well as on two auditory datasets. However, on a third auditory dataset as well as on a visual dataset, GPFA did not show a clear advantage compared to SFA. This matches our experience with other visual datasets (not shown here). We hypothesize that SFA's assumption of the most relevant signals being the slow ones may especially suited for the characteristics of visual data. This also matches the fact that SFA originally was designed for and already proved to work well for signal extraction from visual data sets. A detailed analysis of which algorithm and corresponding measure of predictability is best suited for what kind of data or domain remains a subject of future research.

In practical terms we conclude that GPFA~(2) has some advantages over GPFA~(1). First, its linear time complexity in $k$ (see Section~\ref{sec:time_complexity_matrix_mult}) makes a notable difference in practice (see Section~\ref{sec:experiments:toy}). Second, GPFA~(2) consistently produced better results (see Section~\ref{sec:experiments}) which is a bit surprising given that the fully connected graph of GPFA~(1) is theoretically more sound and also matches the actual evaluation criterion~\eqref{eq:knn_estimate}. Our intuition here is that it is beneficial to give $\vect y_{t+1}$ a central role in the graph because it is a more reliable estimate of the true mean of $p(\vec Y_{t+1} | \vec Y_t = \vect y_t)$ than the empirical mean of all data points (stemming from different distributions) in the fully connected graph.

In the form described above, GPFA performs linear feature extraction. However, we are going to point out three strategies to extend the current algorithm for non-linear feature extraction. The first strategy is very straight-forward and can be applied to the other linear feature extractors as well: In a preprocessing step, the data is expanded in a non-linear way, for instance through all polynomials up to a certain order. Afterwards, application of a linear feature extractor implicitly results in non-linear feature extraction. This strategy is usually applied to SFA, often in combination with hierarchical stacking of SFA nodes which further increases the non-linearities while at the same time regularizing spatially (on visual data)~\citep{Escalante-B.Wiskott-2012a}.

The other two approaches to non-linear feature extraction build upon the graph embedding framework. We already mentioned above that kernel versions of graph embedding are readily available~\citep{YanXuEtAl-2007, CaiHeEtAl-2007}. Another approach to non-linear graph embedding was described for an algorithm called \emph{hierarchical generalized SFA}: A given graph is embedded by first expanding the data in a non-linear way and then calculating a lower-dimensional embedding of the graph on the expanded data. This step is repeated---each time with the original graph---resulting in an embedding for the original graph that is increasingly non-linear with every repetition (see~\citep{Sprekeler-2011} for details).

Regarding the analytical understanding of GPFA, we have shown in Section~\ref{sec:predictive_information} under which assumptions GPFA can be understood as finding the features with the highest predictive information, for instance when the underlying process is assumed to be deterministic but its states disturbed by independent Gaussian noise. If we generally had the goal of minimizing the coding length of the extracted signals (which would correspond to high predictive information) rather than minimizing their next-step variance, then the covariances in GPFA's main objective~\eqref{eq:predictability} needed to be weighted logarithmically. Such an adoption, however, would not be straight forward to include into the graph structure. 

Another information-theoretic concept relevant in this context (besides predictive information) is that of information bottlenecks~\citep{TishbyPereiraEtAl-2000}. Given two random variables $\vec A$ and $\vec B$, an information bottleneck is a compressed variable $\vec T$ that solves the problem $\min_{p(\vect t | \vect a)} I(\vec A; \vec T) - \beta I(\vec T; \vec B)$. Intuitively, $\vec T$ encodes as much information from $\vec A$ about $\vec B$ as possible while being restricted in complexity. When this idea is applied to time series such that $\vec A$ represents the past and $\vec B$ the future, then $\vec T$ can be understood as encoding the most predictable aspects of that time series. In fact, SFA has been shown to implement a special case of such a past-future information bottleneck for Gaussian variables~\citep{CreutzigSprekeler-2008}. The relationship between GPFA and (past-future) information bottlenecks shall be investigated in the future.

In Section~\ref{sec:heuristics} we introduced the heuristic of reducing the variance of the past in addition that of the future. Effectively this groups together parts of the feature space that have similar expected futures. This property may be especially valuable for interactive settings like reinforcement learning. When you consider an agent navigating its environment, it is usually less relevant to know which way it reached a certain state but rather where it can go to from there. That's why state representations encoding the agent's future generalize better and allow for more efficient learning of policies than state representations that encode the agent's past~\citep{LittmanSuttonEtAl-2001,RafolsRingEtAl-2005}. To better address interactive settings, multiple actions may incorporated into GPFA by conditioning the kNN search on actions, for instance. Additional edges in the graph could also allow grouping together features with similar expected rewards. We see such extension of GPFA as an interesting avenue of future research.

\section{Conclusion}
\label{sec:conclusion}

We presented \emph{graph-based predictable feature analysis} (GPFA), a new algorithm for unsupervised learning of predictable features from high-dimensional time series. We proposed to use the variance of the conditional distribution of the next time point given the previous ones to quantify the predictability of the learned representations and showed how this quantity relates to the information-theoretic measure of predictive information. As demonstrated, searching for the projection that minimizes the proposed predictability measure can be reformulated as a problem of graph embedding. Experimentally, GPFA produced very competitive results, especially on auditory STFT datasets, which makes it a promising candidate for every problem of dimensionality reduction (DR) in which the data is inherently embedded in time.


\bibliographystyle{plainnat}
\bibliography{paper}

\end{document}